\documentclass{article}

\usepackage{microtype}
\usepackage{graphicx}
\usepackage{subcaption}
\usepackage{booktabs} 

\usepackage{hyperref}

\usepackage[preprint]{icml2026}

\usepackage{amsmath}
\usepackage{amssymb}
\usepackage{mathtools}
\usepackage{amsthm}
\usepackage{bigstrut}
\usepackage{multirow}
\usepackage{titlesec}
\usepackage{array}
\usepackage{graphicx}
\usepackage{subcaption}
\usepackage{tcolorbox}

\definecolor{SteelBlue}{RGB}{70, 114, 196}
\definecolor{SoftBlue}{RGB}{240, 245, 255}

\titlespacing*{\section}{0pt}{2pt}{1.5pt}

\titlespacing*{\subsection}{0pt}{2pt}{1.5pt}

\titlespacing*{\subsubsection}{0pt}{1.5pt}{1pt}

\usepackage[capitalize,noabbrev]{cleveref}

\theoremstyle{plain}

\theoremstyle{definition}

\theoremstyle{remark}

\usepackage[textsize=tiny]{todonotes}

\begin{document}

\twocolumn[
  \icmltitle{NextMem: Towards Latent Factual Memory for LLM-based Agents}



  \icmlsetsymbol{equal}{*}

  \begin{icmlauthorlist}
    \icmlauthor{Zeyu Zhang}{equal,ruc}
    \icmlauthor{Rui Li}{equal,ruc}
    \icmlauthor{Xiaoyan Zhao}{cuhk}
    \icmlauthor{Yang Zhang}{nus}
    \icmlauthor{Wenjie Wang}{ustc}
    \icmlauthor{Xu Chen}{ruc}
    \icmlauthor{Tat-Seng Chua}{nus}
  \end{icmlauthorlist}

  \icmlaffiliation{ruc}{Renmin University of China}
  \icmlaffiliation{cuhk}{The Chinese University of Hong Kong}
  \icmlaffiliation{nus}{National University of Singapore}
  \icmlaffiliation{ustc}{University of Science and Technology of China}

  \icmlcorrespondingauthor{Yang Zhang}{zhangy@nus.edu.sg}
  \icmlcorrespondingauthor{Xu Chen}{xu.chen@ruc.edu.cn}


  \vskip 0.3in
]



\printAffiliationsAndNotice{}  

\begin{abstract}
Memory is critical for LLM-based agents to preserve past observations for future decision-making, where factual memory serves as its foundational part.
However, existing approaches to constructing factual memory face several limitations.
Textual methods impose heavy context and indexing burdens, while parametric methods suffer from catastrophic forgetting and high costs.
To address these challenges, we introduce NextMem, a latent factual memory framework that utilizes an autoregressive autoencoder to efficiently construct latent memory while ensuring accurate reconstruction.
For better optimization, we propose a two-stage training process, including autoregressive reconstruction alignment and progressive latent substitution.
We also incorporate quantization to reduce storage overhead.
Extensive experiments demonstrate that NextMem achieves superior performance, and excels in retrieval, robustness, and extensibility properties.
We release our code and model checkpoints at \url{https://github.com/nuster1128/NextMem}.
\end{abstract}

\section{Introduction}
\begin{figure}[t]
	\centering
	\begin{subfigure}[t]{\linewidth}
		\centering
		\includegraphics[width=\textwidth]{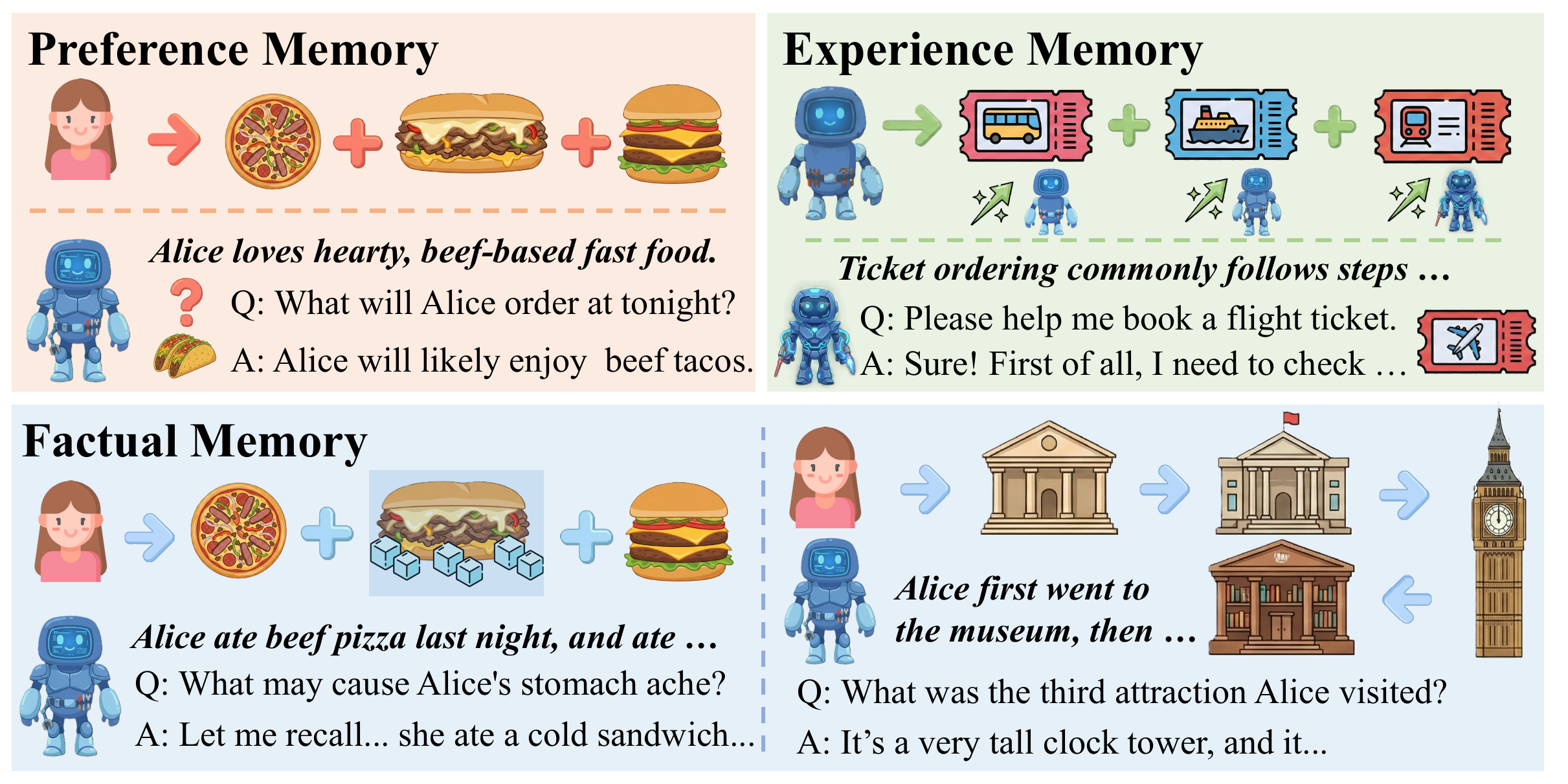}
	\end{subfigure}
	\vspace{-0.6cm}
	\caption{Comparison between task-oriented and factual memory.}
	\label{fig:intro}
	\vspace{-0.8cm}
\end{figure}

\setlength{\abovedisplayskip}{3pt}
\setlength{\belowdisplayskip}{3pt}
In recent years, LLM-based agents have emerged as a new AI paradigm in many fields~\cite{wang2024survey,xi2025rise}, such as personal assistants~\cite{li2024personal} and academic research~\cite{zhang2025deep}.
Memory is among their most critical components, which is responsible for retaining past information to support future decision-making~\cite{zhang2025survey}.
Although the information is typically stored as multiple levels of memory~\cite{li2025memos}, \textbf{factual memory} still remains as their foundation, preserving details of observed facts.
Compared to other task-oriented memories, such as preference~\cite{sun2025preference} and experience~\cite{zhao2024expel}, which commonly require task-specific extraction from the original information, factual memory emphasizes its lossless preservation, as we present their comparison in \textbf{Figure~\ref{fig:intro}}.

Previous research typically represents memory in two forms, including texts and parameters~\cite{zhang2025survey}.
Textual memory is commonly utilized as a context for prompting LLMs, which can be stored and retrieved by databases~\cite{park2023generative, zhong2024memorybank}.
However, it increases context length and indexing overhead when storing and using a large amount of detailed facts.
Other studies incorporate information by modifying the parameters of LLMs~\cite{han2024parameter, zhang2024comprehensive}.
However, they often suffer from catastrophic forgetting and high costs to store detailed facts accurately.
Both paradigms have obvious limitations in managing factual memory.
Recent studies in LLM reasoning propose to leverage latent embeddings to represent intermediate reasoning steps~\cite{hao2024training, xu2025softcot}, as well as compresses task instructions into latent embeddings for optimization~\cite{mu2023learning, wu2025tokmem}.
These studies show the potential of latent representations in storing information.
Nevertheless, they primarily focus on improving reasoning efficiency or task adaptation, rather than preserving factual information.

Motivated by these studies, we intend to leverage latent representations to address the limitations in managing factual memory.
Our primary goals are two-fold:
(1) Efficiently transform textual memory into shorter latent representations compatible with LLMs.
(2) Latent representations can be accurately and efficiently reconstructed into the original memory.
We emphasize reconstruction because factual memory requires lossless preservation rather than partial extraction. Therefore, unlike extraction or indexing approaches, our encoding and decoding processes should be reversible.

In this study, we propose a simple yet effective latent memory framework, named \textbf{NextMem}.
We design an autoregressive autoencoder built upon LLMs for efficient memory encoding and decoding.
Our training process has two critical stages, including autoregressive reconstruction alignment and progressive latent substitution, with further quantization to reduce storage cost while preserving accuracy.
Additionally, we conduct extensive experiments to verify model effectiveness and reveal key properties and insights.
Our experiments also indicate that the latent memory encoded by NextMem exhibits superior retrieval properties, robustness, and extensibility.
To facilitate future research, we have released our code and model checkpoints at \url{https://github.com/nuster1128/NextMem}.

Our contributions are summarized as follows: \\
$\bullet$ We introduce a simple yet effective framework for latent factual memory, with autoregressive reconstruction alignment and progressive latent substitution.\\
$\bullet$ We integrate quantization methods into the latent memory of our framework, which reduces storage overhead while maintaining competitive performance. \\
$\bullet$ We validate our approach through extensive experiments and provide further insights. We provide the research community with open-source code and model checkpoints.


\section{Preliminaries}
\label{sec:preliminaries}

\subsection{Memory in LLM-based Agents}
Unlike vanilla LLMs that generate single-turn responses, LLM-based agents interact iteratively with their environments.
We formalize this process as a Markov Decision Process (MDP)~\cite{sutton1998reinforcement}.
Specifically, let $S$ and $A$ denote the state and action spaces.
The environment is featured by a transition function $E: S \times A \rightarrow S$ with rewards.
At each timestep $t$, the agent selects an action $a^t = f(s^t)$ based on the current state $s^t \in S$ via a policy $f$.
Then, the state will be updated by the environment $s^{t+1} = E(s^t, a^t)$ with a reward $r^t$, which can be further observed by agents to take the next action $a^{t+1}$.
This process repeats until task completion, and the objective is to optimize the policy $f$ to maximize the cumulative reward $\sum r^t$.

Memory is fundamental to agent decision-making. To accommodate diverse memory representations, we conceptualize the memory procedure as an encoding-decoding process rather than the traditional paradigm of storage, retrieval, and utilization~\cite{zhang2025learn}.
Under this framework, an agent \textbf{encodes} historical information into memory, which is subsequently \textbf{decoded} to augment LLM inference.
Specifically, let $M^t$ denote the memory state at timestep $t$. The encoding process is formalized as:
$$
M^{t} = \text{Encode}(s^t, M^{t-1}),
$$
which is executed upon observing the current state $s^t$.
The decoding process is integrated into agent's policy function:
$$
f(s^t) = \text{LLM} \circ \text{Decode}(s^t, M^{t}),
$$
where function $\text{LLM}$ represents LLM inference with core capabilities, such as reasoning, inference, and tool calling.
Importantly, information encoded into $M^t$ at step $t$ may not be utilized until a later timestep, reflecting the asynchronous feature of encoding and decoding in agentic workflows.

\subsection{Representation of Agent Memory}
Memory can be manifested in different representations, and our encoding-decoding framework provides a unified perspective for these memory forms as follows.

\textbf{Textual Memory.}
Textual memory stores information in text format and leverages in-context learning to integrate relevant data into LLM inference~\cite{zhang2025survey}.
This approach often employs indexing structures for query-based retrieval.
The encoding process is formulated as:
\begin{gather*}
	m^t = \text{LLM}(\theta; p_{\text{extract}} \Vert s^t) \in \mathcal{V}^n, \\
	M^{t} = M^{t-1} \oplus \{(m^t, \text{Index}(m^t))\},
\end{gather*}
where $p_{\text{extract}}$ denotes the prompt used to extract memory from observations, $\Vert$ represents concatenation, and $\oplus$ signifies a structural merge.
Here, we use $\text{LLM}(\theta;\cdot)$ to represent LLM inference parameterized by $\theta$ over a vocabulary $\mathcal{V}$ with sequence length $n$, and use $\text{Index}(\cdot)$ to denote the approach of indexing.
Then, the decoding process can be defined as:
\begin{gather*}
	q^t = \text{LLM}(\theta; p_{\text{intent}} \Vert s^t) \in \mathcal{V}^n, \\
	c^t = \text{Retrieval}(\text{Index}(q^t), M^t), \\
	f(s^t) = \text{LLM}(\theta; p_{\text{instruct}} \Vert s^t \Vert c^t),
\end{gather*}
where $p_{\text{intent}}$ is the prompt for generating the retrieval intent, $p_{\text{instruct}}$ is the task-specific instruction prompt, and $\text{Retrieval}(\cdot)$ denotes the operation of fetching relevant context $c^t$ based on the established indexes.

\textbf{Parametric Memory.}
Parametric memory incorporates new information by modifying model parameters, which are subsequently integrated with the base parameters at the inference stage.
The encoding process can be formalized as:
\begin{gather*}
	\Delta \theta^t = h_{\text{meta}}(s^t, \theta^0 \oplus \Delta \theta^{t-1}) \in \mathbb{R}^{d},\\
	M^t := \theta^0 \oplus \Delta \theta^t,
\end{gather*}
where $\theta^0$ denotes the base parameters of the LLM. The function $h_{\text{meta}}(\cdot)$ is designed to predict parameter modifications within the space $\mathbb{R}^{d}$ based on the current observation $s^t$.
The decoding process is defined as:
\begin{gather*}
	f(s^t) = \text{LLM}(\theta^0 + \Delta \theta^t; p_{\text{instruct}} \Vert s^t),
\end{gather*}
where $\Delta \theta^t$ represents the cumulative parameter modifications that encapsulate the historical information for the task.

\begin{figure*}[t]
	\centering
	\begin{subfigure}[t]{\linewidth}
		\centering
		\includegraphics[width=\textwidth]{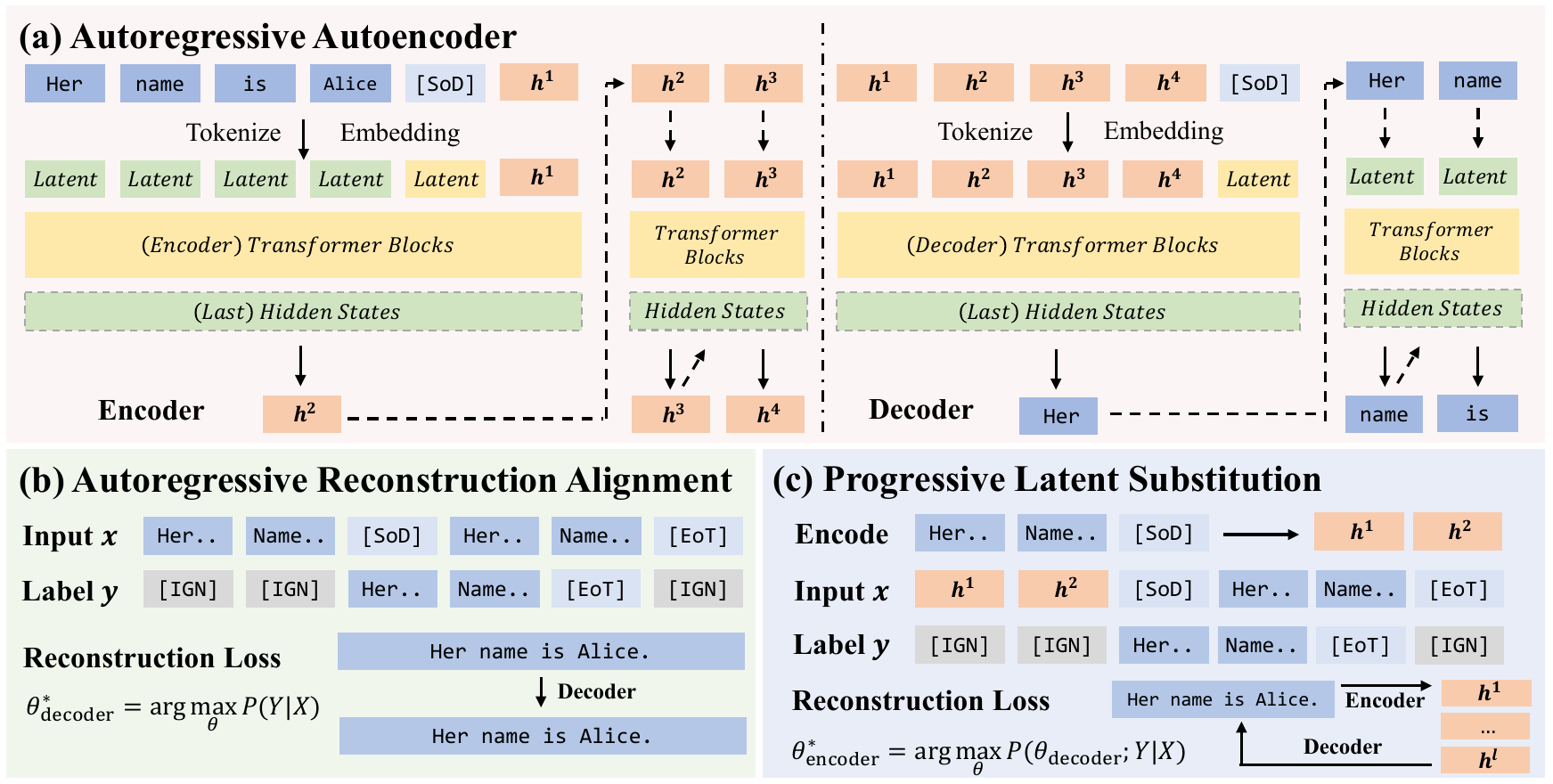}
	\end{subfigure}
	\vspace{-0.6cm}
	\caption{Overview of NextMem framework.}
	\label{fig:method}
	\vspace{-0.6cm}
\end{figure*}

\textbf{Latent Memory.}
Beyond the above paradigms, latent memory offers a distinct approach by using an encoder to transform textual information into latent representations, which enables reducing sequence length and inference latency.
Its encoding process is formalized as:
\begin{gather*}
	m^t = h_{\text{encode}}(s^t) \in \mathbb{R}^{L \times d}, \\
	M^t = M^{t-1} \oplus \{(m^t, \text{Index}(m^t))\},
\end{gather*}
where $h_{\text{encode}}$ transforms information from the textual space $V^n$ into a latent space $\mathbb{R}^{L \times d}$, with $L \ll n$. For the decoding procedure, we have the following definition:
\begin{gather*}
	q^t = h_{\text{encode}}(s^t) \in \mathbb{R}^{L \times d}, \\
	c^t = \text{Retrieval}(\text{Index}(q^t), M^t), \\
	f(s^t) = \text{LLM}(\theta; p_{\text{instruct}} \Vert s^t, c^t), 
\end{gather*}
where the retrieved latent context $c^t$ can be integrated into the input embeddings for LLM inference.

\section{Methods}
\label{sec:methods}

\subsection{Overview}
In this paper, we propose a simple yet effective autoregressive autoencoder to generate latent factual memory, as presented in \textbf{Figure~\ref{fig:method}(a)}.
Our model can forwardly transform textual information into latent representations that are compatible with LLMs' inputs.
It can be accurately decoded back into the original text to ensure fine-grained preservation.
We design a two-stage training procedure to establish text-to-text, latent-to-text, and text-to-latent information transformation, which includes autoregressive reconstruction alignment and progressive latent substitution.

\subsection{Autoregressive Autoencoder}
Our autoencoder is built on a causal language model architecture. It can be decomposed into three primary parts: \\
\noindent \textbf{(1) Embedding Layer} ($h_{\text{emb}}$): It maps discrete input tokens $s \in \mathcal{V}^n$ into a continuous embedding space $\mathbb{R}^{n \times d}$. \\
\noindent \textbf{(2) Transformer Blocks} ($h_{\text{trans}}$): It consists of a stack of Transformer decoder layers~\cite{vaswani2017attention}. Each layer integrates multi-head self-attention and a feed-forward network, with residual connections and normalization. \\
\noindent \textbf{(3) Language Modeling Head} ($h_{\text{lmh}}$): It projects the final hidden states into a probability distribution over the vocabulary space for next-token predictions.

To improve model efficiency, the encoder and decoder share an identical architecture with two distinct weight sets $\theta_{\text{encode}}$ and $\theta_{\text{decode}}$.
Furthermore, we introduce a special token \texttt{[SoD]} to signify the start of transformation.
During the encoding phase, we append it to the original input sequence $s = [s_1, s_2, \dots, s_n] \in \mathcal{V}^n$ before mapping to the initial input embedding:
\begin{gather*}
	\mathbf{E}^0 = h_{\text{emb}}(\theta_{\text{encode}}; s \Vert \texttt{[SoD]}) \in \mathbb{R}^{(n+1) \times d}.
\end{gather*}
The embedding is then processed through the Transformer blocks. We extract the hidden state from the last position of the final layer as the first latent embedding:
\begin{gather*}
	\mathbf{h}^1 = h_{\text{trans}}(\theta_{\text{encode}}; \mathbf{E}^0)_{(T, n+1)} \in \mathbb{R}^d,
\end{gather*}
where $T$ denotes the total number of Transformer blocks.
We iteratively add the previously obtained latent embedding to the input embeddings.
The $i$-th step is defined as:
\begin{gather*}
	\mathbf{E}^i = [\mathbf{E}^{i-1}; \mathbf{h}^{i}], \\
	\mathbf{h}^{i+1} = h_{\text{trans}}(\theta_{\text{encode}}; \mathbf{E}^{i})_{(T, n+i+1)}.
\end{gather*}
The final latent representation is the concatenation of all $l$ generated latent embeddings:
\begin{gather*}
	\mathbf{H}^{(l)} = [\mathbf{h}^1; \mathbf{h}^2; \dots; \mathbf{h}^l] \in \mathbb{R}^{l \times d}.
\end{gather*}
In the decoding phase, we first map the instruction suffix $p_{\text{suffix}}$ (e.g., \texttt{[SoD]}) into the input embedding space:
\begin{gather*}
	\mathbf{E}_{\text{suffix}} = h_{\text{emb}}(\theta_{\text{decode}}; p_{\text{suffix}}).
\end{gather*}
The input of the decoder is then formed by concatenating the latent representation $\mathbf{H}^{(l)} $ with this suffix embedding:
\begin{gather*}
	\mathbf{E}_{\text{input}} = [ \mathbf{H}^{(l)} ; \mathbf{E}_{\text{suffix}} ].
\end{gather*}
The probability distribution over the vocabulary $\mathcal{V}$ for the next-token prediction is computed as:
\begin{gather*}
	p(\mathcal{V} \mid \mathbf{E}_{\text{input}}) = h_{\text{lmh}} \circ h_{\text{trans}}(\theta_{\text{decode}}; \mathbf{E}_{\text{input}}).
\end{gather*}
Based on this distribution, the decoder samples tokens and generates the output sequence $o = [o_1, o_2, \dots, o_m] \in \mathcal{V}^m$ in an autoregressive manner.

\subsection{Autoregressive Reconstruction Alignment}
To optimize our autoencoder, we propose a two-stage training procedure.
In the first stage, autoregressive reconstruction alignment is designed to enable the model to transform information from textual space to textual space autoregressively, as shown in \textbf{Figure~\ref{fig:method}(b)}.
Our training samples are constructed in a self-supervised manner.
Specifically, for each original sequence $s = [s_1, s_2, \dots, s_n]$, we define the input sequence $x$ and the corresponding target labels $y$ as:
\begin{align*}
	x &= [ s_1, s_2, \dots, s_n, \texttt{[SoD]}, s_1, s_2, \dots, s_{n} ], \\
	y &= 
	\left[
	\underbrace{
		\texttt{[IGN]}, ...}_{n}, s_1, ..., s_n, \texttt{[EoT]}
	\right],
\end{align*}
where $\texttt{[IGN]}$ denotes positions where the loss calculation is ignored, and $\texttt{[EoT]}$ is a pre-trained token signifying the end of the text.
We then fine-tune the causal language model by maximizing the following likelihood:
\begin{gather*}
	\theta_{\text{decode}}^* = \arg\max_{\theta_{\text{decode}}}P(\theta_{\text{decode}}; Y | X),
\end{gather*}
where $P(\theta; Y|X) = h_{\text{lmh}} \circ h_{\text{trans}} \circ h_{\text{emb}}(\theta; X)$.

\subsection{Progressive Latent Substitution}
Following the initial alignment, we introduce progressive latent substitution, which further enables the encoder to transform information into latent representations and reconstruct it.
As illustrated in \textbf{Figure~\ref{fig:method}(c)}, it comprises $L$ progressive steps.
At the $k$-th step, for an original sequence $s$, we first generate a $k$-length latent representation $\mathbf{H}^{(k)}$ via the encoding process.
Then, we substitute the first $k$ blocks of the original sequence (each of block size $B$) with the latent representations. The remaining textual sequence is
\begin{gather*}
	\tilde{s}^{(k)} = [ s_{k \cdot B + 1}, s_{k \cdot B + 2}, \dots, s_n ].
\end{gather*}
We then construct the input embedding as follows:
\begin{gather*}
	\mathbf{x}^{(k)}(\theta; s) = \left[ \mathbf{H}^{(k)}; h_{\text{emb}}(\theta; \tilde{s}^{(k)} \Vert \texttt{[SoD]} \Vert s) \right].
\end{gather*}
Compared to the last alignment phase, we substitute several blocks of the original text with their corresponding latent representations $\mathbf{H}^{(k)}$.
This forces the model to rely on latent representations to recover the missing textual information.
The target label for the decoder is defined as:
\begin{gather*}
	y^{(k)} = 
	\left[
	\underbrace{
		\texttt{$\texttt{[IGN]}$}, ...}_{n}, s_1, ..., s_{k \cdot B},
	\underbrace{
		\texttt{[IGN]}, ...}_{n - k \cdot B}, \texttt{[EoT]}
	\right].
\end{gather*}
In this stage, only the encoder parameters $\theta_{\text{encode}}$ are optimized, while keeping the decoder parameters $\theta_{\text{decode}}$ frozen. This ensures that the encoder learns to produce representations compatible with the pre-aligned decoder.
The optimization objective is formulated as:
\begin{gather*}
	\theta_{\text{encode}}^{(k)} = \arg\max_{\theta_{\text{encode}}}P\left( \theta_{\text{decode}}; Y^{(k)} | \mathbf{X}^{(k)} \left( \theta_{{\text{encode}}}; S \right) \right),
\end{gather*}
where $P(\theta_{\text{decode}}; Y^{(k)}|\mathbf{X}^{(k)}) = h_{\text{lmh}} \circ h_{\text{trans}}(\theta_{\text{decode}}; \mathbf{X}^{(k)})$, and $S$ denotes the original text set of training corpus.

For each step $k$, the encoder parameters are initialized with the weights from the previous step, i.e., $\theta_{{\text{encode}}}^{(k)} \leftarrow \theta_{\text{encode}}^{(k-1)}$. For the first step, we initialize the encoder using the pre-aligned decoder weights $\theta_{\text{decode}}^*$. The final encoder parameters are denoted as $\theta_{\text{encode}}^* = \theta_{\text{encode}}^{(L)}$ after $L$ steps of progressive substitution.
To enhance optimization efficiency and stability, we apply a stop-gradient operation to the hidden state $\mathbf{h}^{i-1}$ when computing the gradient $\nabla_\theta \mathbf{h}^{i}$.
This detachment prevents the backpropagation of gradients through multiple recurrence, significantly reducing the computational cost.
We employ LoRA~\cite{hu2022lora} to implement both the encoder and decoder.
This approach allows us to switch between the encoder and decoder by simply swapping their LoRA adapters while sharing the same backbone, which avoids the replicated model loading.

\subsection{Latent Memory Quantization}
After the two-stage training, we observe that the latent representations exhibit strong robustness (see \textbf{Section~\ref{sec:exp_robustness}}).
To further reduce storage overhead, we employ 4-bit NormalFloat (NF4) quantization~\cite{dettmers2023qlora} for further compression.
The codebook is defined as a fixed set of NF4 values $\mathcal{C} = \{c_0, c_1, \dots, c_{15}\}$. 
The quantization process maps the high-precision latent representation $\mathbf{H}^{(L)}$ to 4-bit indices $\mathbf{Q}^{(L)}$ and an associated scale vector $\mathbf{s}$.
First, we compute the quantization scale $\mathbf{s} \in \mathbb{R}^d$ for each feature dimension $j$ by determining the maximum absolute value:
\begin{gather*}
	\mathbf{s}_j = \max_{1 \le i \le l} |\mathbf{H}_{i,j}^{(L)}|.
\end{gather*}
The input is then normalized element-wise using
\begin{gather*}
	\tilde{\mathbf{H}}_{i,j}^{(L)} = \frac{\mathbf{H}_{i,j}^{(L)}}{\mathbf{s}_j + \epsilon},
\end{gather*}
where $\epsilon$ is a small constant for numerical stability. Subsequently, each normalized element is mapped to its nearest centroid in the codebook $\mathcal{C}$ via:
\begin{gather*}
	\mathbf{Q}_{i,j}^{(L)} = \arg\min_{k} \left| \tilde{\mathbf{H}}_{i,j}^{(L)} - c_k \right|.
\end{gather*}
The resulting indices $\mathbf{Q}^{(L)}$ are stored as 4-bit unsigned integers, while the scales $\mathbf{s}$ are cast to the FP8 format (specifically \texttt{float8\_e4m3fn}) to further optimize memory efficiency.
To reconstruct the approximation $\hat{\mathbf{H}}^{(L)}$, we retrieve the codebook values with their indices and rescale them.
Actually, we have explored more methods for higher sparsity, but they failed to maintain sufficient reconstruction accuracy.
More details are provided in \textbf{Appendix~\ref{appendix:failure_case}}.

\subsection{Efficiency and Scalability Analysis}
NextMem improves inference efficiency by compressing extensive textual observations into compact latent representations.
It alleviates context window pressure for agents to allocate more token capacity to complex reasoning and long-horizon planning.
Additionally, the adoption of shared backbone parameters with LoRA adapters and NF4 quantization minimizes the model’s memory overhead while enabling high-density storage of factual records.
These optimizations collectively facilitate the scalable deployment of factual memory within resource-constrained agentic workflows.

\section{Experiments}
\label{sec:experiments}

\begin{table*}[t]
	\centering
	\caption{Performance comparison of factual reconstruction (\textit{i.e.,} task 1 for memory storage) across multiple datasets. In each assessment, values in \textbf{bold} represent the best results, and those with an \underline{underline} represent the second-best results.}
	\vspace{-0.15cm}
	\begin{tabular}{cccccccc}
		\hline
		\hline
		\textbf{Datasets} & \textbf{Methods} & \textbf{F1} & \textbf{ROUGE-1} & \textbf{ROUGE-L} & \textbf{METEOR} & \textbf{BLEU} & \textbf{BertScore} \bigstrut\\
		\hline
		\multicolumn{1}{c}{\multirow{5}[2]{*}{HotpotQA}} & DyPRAG & 0.0305  & 0.0347  & 0.0337  & 0.0187  & 0.0000  & 0.7983  \bigstrut[t]\\
		& DeepSeek-OCR & 0.4540  & 0.5492  & 0.5374  & 0.3987  & 0.2432  & 0.8664  \\
		& ICAE  & 0.7890  & 0.8570  & 0.8340  & 0.7493  & 0.5782  & 0.9581  \\
		& NextMem-Dense & \textbf{0.9820} & \textbf{0.9862} & \textbf{0.9854} & \textbf{0.9820} & \textbf{0.9633} & \textbf{0.9966} \\
		& NextMem-Sparse & \underline{0.9805}  & \underline{0.9842}  & \underline{0.9833}  & \underline{0.9810}  & \underline{0.9612}  & \underline{0.9962}  \bigstrut[b]\\
		\hline
		\multicolumn{1}{c}{\multirow{5}[2]{*}{RACE}} & DyPRAG & 0.0696  & 0.0826  & 0.0689  & 0.0341  & 0.0000  & 0.8158  \bigstrut[t]\\
		& DeepSeek-OCR & 0.4068  & 0.4509  & 0.4268  & 0.3626  & 0.2371  & 0.8481  \\
		& ICAE  & 0.6077  & 0.6775  & 0.6117  & 0.5555  & 0.3503  & 0.9370  \\
		& NextMem-Dense & \underline{0.8552}  & \textbf{0.8838} & \underline{0.8580}  & \underline{0.8691}  & \textbf{0.6995} & \textbf{0.9735} \\
		& NextMem-Sparse & \textbf{0.8554} & \underline{0.8833}  & \textbf{0.8583} & \textbf{0.8705} & \underline{0.6975}  & \underline{0.9731}  \bigstrut[b]\\
		\hline
		\multicolumn{1}{c}{\multirow{5}[2]{*}{SQuAD}} & DyPRAG & 0.0493  & 0.0510  & 0.0477  & 0.0221  & 0.0000  & 0.8040  \bigstrut[t]\\
		& DeepSeek-OCR & 0.3657  & 0.4018  & 0.3755  & 0.3169  & 0.1864  & 0.8289  \\
		& ICAE  & 0.7084  & 0.7709  & 0.7163  & 0.6508  & 0.4501  & 0.9536  \\
		& NextMem-Dense & \textbf{0.8920} & \textbf{0.9128} & \textbf{0.8886} & \textbf{0.8958} & \textbf{0.7581} & \textbf{0.9785} \\
		& NextMem-Sparse & \underline{0.8860}  & \underline{0.9069}  & \underline{0.8826}  & \underline{0.8897}  & \underline{0.7443}  & \underline{0.9778}  \bigstrut[b]\\
		\hline
		\multicolumn{1}{c}{\multirow{5}[2]{*}{LoCoMo}} & DyPRAG & 0.0901  & 0.1143  & 0.0932  & 0.0501  & 0.0000  & 0.8335  \bigstrut[t]\\
		& DeepSeek-OCR & 0.5179  & 0.6421  & 0.6272  & 0.4627  & 0.3139  & 0.8962  \\
		& ICAE  & 0.6986  & 0.7815  & 0.7515  & 0.7043  & 0.4730  & 0.9560  \\
		& NextMem-Dense & \underline{0.9611}  & \textbf{0.9742} & \underline{0.9704}  & \textbf{0.9640} & \underline{0.9063}  & \textbf{0.9946} \\
		& NextMem-Sparse & \textbf{0.9615} & \underline{0.9741}  & \textbf{0.9705} & \underline{0.9637}  & \textbf{0.9070} & \underline{0.9944}  \bigstrut[b]\\
		\hline
		\multicolumn{1}{c}{\multirow{5}[2]{*}{LongMemEval}} & DyPRAG & 0.1338  & 0.1643  & 0.1331  & 0.0643  & 0.0000  & 0.8360  \bigstrut[t]\\
		& DeepSeek-OCR & 0.4685  & 0.5375  & 0.5106  & 0.4116  & 0.2713  & 0.8681  \\
		& ICAE  & 0.7015  & 0.7510  & 0.7007  & 0.6634  & 0.4690  & 0.9535  \\
		& NextMem-Dense & \textbf{0.9436} & \textbf{0.9620} & \textbf{0.9555} & \textbf{0.9466} & \textbf{0.8784} & \textbf{0.9905} \\
		& NextMem-Sparse & \underline{0.9362}  & \underline{0.9554}  & \underline{0.9486}  & \underline{0.9397}  & \underline{0.8692}  & \underline{0.9891}  \bigstrut[b]\\
		\hline
		\hline
	\end{tabular}%
	\vspace{-0.55cm}
	\label{tab:result_task1}%
\end{table*}%

\subsection{Experimental Settings}
\label{sec:exp_setting}
To validate the effectiveness of our model, we conduct extensive experiments and analyze its results from multiple aspects.
Our primary evaluation involves three tasks that are closely related to agent memory, including (1) \textbf{Factual Reconstruction}, (2) \textbf{Contextual Generation}, and (3) \textbf{Dense Passage Retrieval}.
These three tasks correspond to memory storage, utilization, and retrieval, respectively.
In addition, we further explore compression ratio influence, robustness, forgetting effects and other features of our approaches.

The datasets utilized in our experiments include: \\
$\bullet$ \textbf{SQuAD}~\cite{rajpurkar2016squad}: A reading comprehension dataset that requires to answer questions according to passages extracted from Wikipedia. \\
$\bullet$ \textbf{HotpotQA}~\cite{yang2018hotpotqa}: A question answering dataset that requires multi-hop reasoning based on information across multiple Wikipedia documents.\\
$\bullet$ \textbf{RACE}~\cite{lai2017race}: A reading comprehension dataset from English exams that test reasoning and understanding. \\
$\bullet$ \textbf{LoCoMo}~\cite{maharana2024evaluating}: A simulated dataset for evaluating long-term memory of LLM-based agents through multi-session conversations among different speakers.\\
$\bullet$ \textbf{LongMemEval}~\cite{wu2024longmemeval}: A dataset that is constructed to evaluate long-term memory capabilities of personal assistants in the scenario of user-agent interactions.

The major baselines in our experiments include: \\
$\bullet$ \textbf{DeepSeek-OCR}~\cite{wei2025deepseek}: A vision-language model that utilizes context optical compression to compress texts into images before performing inference by LLMs.
To unify the scale of the hidden representations, we employ the 240 $\times$ 240 pixel (16 latent tokens) patterns.
\\
$\bullet$ \textbf{ICAE}~\cite{ge2023context}: A context compression model that can convert paragraphs into memory slots for LLM inference, which is based on optimizable memory tokens.
We employ the shortest public checkpoint with 128 tokens.
\\
$\bullet$ \textbf{DyPRAG}~\cite{tan2025dynamic}: A framework that converts paragraphs into parametric knowledge (specifically LoRA adapters) at test time using a lightweight hypernetwork.

\begin{table*}[t]
	\centering
	\caption{Performance comparison of contextual generation (\textit{i.e.,} task 2 for memory utilization) across various datasets. For all non-oracle methods, the best results are highlighted in \textbf{bold}, and the second-best results are \underline{underlined}.}
	\vspace{-0.1cm}
	\begin{tabular}{ccccccccc}
		\hline
		\hline
		\multirow{2}[4]{*}{\textbf{Methods}} & \multicolumn{2}{c}{\textbf{HotpotQA}} & \multicolumn{2}{c}{\textbf{SQuAD}} & \multicolumn{2}{c}{\textbf{LoCoMo}} & \multicolumn{2}{c}{\textbf{LongMemEval}} \bigstrut\\
		\cline{2-9}          & \textbf{Comp.} & \textbf{DeComp.} & \textbf{Comp.} & \textbf{DeComp.} & \textbf{Comp.} & \textbf{DeComp.} & \textbf{Comp.} & \textbf{DeComp.} \bigstrut\\
		\hline
		DyPRAG & 0.5000  & 0.3789  & 0.2659  & 0.2023  & 0.0191  & 0.0239  & 0.0800  & 0.0971  \bigstrut[t]\\
		DeepSeek-OCR & \underline{0.5673}  & 0.3744  & 0.2124  & 0.2225  & 0.0766  & 0.1435  & 0.1943  & 0.1543  \\
		ICAE  & \textbf{0.8565} & \textbf{0.8229} & \textbf{0.7775} & 0.7066  & \textbf{0.5407} & 0.5215  & \textbf{0.4971} & 0.5029  \\
		NextMem-Dense & 0.5179  & 0.8072  & \underline{0.3223}  & \underline{0.7572}  & \underline{0.2871}  & \textbf{0.5407} & 0.1971  & \underline{0.5400}  \\
		NextMem-Sparse & 0.4978  & \underline{0.8184}  & 0.3092  & \textbf{0.7630} & 0.2679  & \underline{0.5263}  & \underline{0.2029}  & \textbf{0.5486} \\
		\textcolor[rgb]{ .502,  .502,  .502}{*Oracle} & \textcolor[rgb]{ .502,  .502,  .502}{---} & \textcolor[rgb]{ .502,  .502,  .502}{0.9350} & \textcolor[rgb]{ .502,  .502,  .502}{---} & \textcolor[rgb]{ .502,  .502,  .502}{0.9335} & \textcolor[rgb]{ .502,  .502,  .502}{---} & \textcolor[rgb]{ .502,  .502,  .502}{0.6986} & \textcolor[rgb]{ .502,  .502,  .502}{---} & \textcolor[rgb]{ .502,  .502,  .502}{0.6971} \bigstrut[b]\\
		\hline
		\hline
	\end{tabular}%
	\label{tab:result_task2}%
	\vspace{-0.05cm}
\end{table*}%

Our models are denoted as \textbf{NextMem-Dense} for the dense version and \textbf{NextMem-Sparse} with quantization, both generating 15 latent tokens.
Besides, we also provide special models to facilitate comparison, including \textbf{Textual Memory}~\cite{zhang2025memengine}, and \textbf{BGE}~\cite{chen2024bge}.

As for the metrics in our experiment, we utilize \textbf{F1 Score}~\cite{rajpurkar2016squad}, \textbf{ROUGE-1}, \textbf{ROUGE-L}~\cite{lin2004rouge}, \textbf{METEOR}~\cite{banerjee2005meteor}, \textbf{BLEU}~\cite{papineni2002bleu}, \textbf{BertSore}~\cite{zhang2019bertscore} for the factual reconstruction task.
Besides, we adopt \textbf{Accuracy} by LLM-as-Judge~\cite{gu2024survey} for the contextual generation task, and use \textbf{Hit@5}, \textbf{Recall@5}, \textbf{MRR@5}, \textbf{MAP@5}, \textbf{DCG@5}, \textbf{NDCG@5}~\cite{schutze2008introduction, karpukhin2020dense} for the dense passage retrieval task.

For common settings, we employ Qwen3-8B~\cite{yang2025qwen3} as the primary backbone except for specific checkpoint requirements.
We set the chunk size of textual references as 128.
Due to the page limitation, we place the full details of reproduction in \textbf{Appendix~\ref{appendix:reproduction}}, influence of block size in \textbf{Appendix~\ref{appendix:exp_block_mask_size}}, forgetting effects in \textbf{Appendix~\ref{appendix:forgetting_effect}}, and case studies in \textbf{Appendix~\ref{appendix:case_study}}.

\subsection{Major Performances}
\label{sec:exp_major}
In order to provide a more comprehensive evaluation, our major experiments include three diverse tasks, corresponding to the storage, utilization, and retrieval of latent memory.

\textbf{Task 1: Factual Reconstruction (Memory Storage)}\\
Factual memory relies on the precision of details. Therefore, we measure the ability of baselines in information reconstruction.
Specifically, we extract reference paragraphs from datasets and apply sentence sampling to improve data diversity and uniformity.
In this task, all the baselines are required to encode texts to latent representations and decode them back to the original form before calculating their consistency.
The results are presented in \textbf{Table~\ref{tab:result_task1}}.

The results show that our proposed methods significantly outperform other baselines across various datasets.
Specifically, NextMem-Dense achieves the highest scores in most scenarios, substantially exceeding ICAE.
In addition, NextMem-Sparse maintains highly competitive performance, which shows the effectiveness of our quantization strategy.
In contrast, previous methods such as DyPRAG and DeepSeek-OCR exhibit limited reconstruction capabilities.
These results validate the superior ability of NextMem to achieve high-fidelity storage of latent memory.

\begin{table*}[t]
	\centering
	\caption{Performance comparison of dense passage retrieval (task 3 for memory retrieval) across various datasets. For all reconstruction models, the best results are highlighted in \textbf{bold}, and the second-best results are \underline{underlined}.}
	\vspace{-0.1cm}
	\begin{tabular}{cccccccc}
		\hline
		\hline
		\textbf{Datasets} & \textbf{Methods} & \textbf{Hit@5} & \textbf{Recall@5} & \textbf{MRR@5} & \textbf{MAP@5} & \textbf{DCG@5} & \textbf{NDCG@5} \bigstrut\\
		\hline
		\multicolumn{1}{c}{\multirow{5}[2]{*}{HotpotQA}} & DeepSeek-OCR & 0.3358  & 0.1487  & 0.1659  & 0.0730  & 0.2260  & 0.1171  \bigstrut[t]\\
		& ICAE  & 0.4453  & 0.2217  & 0.3187  & 0.1555  & 0.4058  & 0.2126  \\
		& NextMem-Dense & \textbf{0.7245} & \underline{0.3925}  & \textbf{0.5194} & \textbf{0.2793} & \underline{0.6673}  & \underline{0.3680}  \\
		& NextMem-Sparse & \underline{0.7208}  & \textbf{0.4030} & \underline{0.5107}  & \underline{0.2788}  & \textbf{0.6683} & \textbf{0.3687} \\
		& \textcolor[rgb]{ .502,  .502,  .502}{*BGE} & \textcolor[rgb]{ .502,  .502,  .502}{0.9585} & \textcolor[rgb]{ .502,  .502,  .502}{0.6681} & \textcolor[rgb]{ .502,  .502,  .502}{0.8063} & \textcolor[rgb]{ .502,  .502,  .502}{0.5442} & \textcolor[rgb]{ .502,  .502,  .502}{1.1756} & \textcolor[rgb]{ .502,  .502,  .502}{0.6438} \bigstrut[b]\\
		\hline
		\multicolumn{1}{c}{\multirow{5}[2]{*}{LoCoMo}} & DeepSeek-OCR & 0.0676  & 0.0321  & 0.0269  & 0.0115  & 0.0383  & 0.0206  \bigstrut[t]\\
		& ICAE  & 0.1210  & 0.0530  & 0.0577  & 0.0254  & 0.0789  & 0.0411  \\
		& NextMem-Dense & \textbf{0.4377} & \textbf{0.2132} & \textbf{0.2418} & \textbf{0.1183} & \textbf{0.3455} & \textbf{0.1768} \\
		& NextMem-Sparse & \underline{0.4342}  & \underline{0.2087}  & \underline{0.2310}  & \underline{0.1111}  & \underline{0.3304}  & \underline{0.1692}  \\
		& \textcolor[rgb]{ .502,  .502,  .502}{*BGE} & \textcolor[rgb]{ .502,  .502,  .502}{0.8007} & \textcolor[rgb]{ .502,  .502,  .502}{0.4824} & \textcolor[rgb]{ .502,  .502,  .502}{0.5061} & \textcolor[rgb]{ .502,  .502,  .502}{0.3181} & \textcolor[rgb]{ .502,  .502,  .502}{0.7953} & \textcolor[rgb]{ .502,  .502,  .502}{0.4166} \bigstrut[b]\\
		\hline
		\multicolumn{1}{c}{\multirow{5}[2]{*}{LongMemEval}} & DeepSeek-OCR & 0.4200  & 0.3125  & 0.1528  & 0.1133  & 0.2315  & 0.1747  \bigstrut[t]\\
		& ICAE  & 0.5480  & 0.4169  & 0.2437  & 0.1840  & 0.3606  & 0.2596  \\
		& NextMem-Dense & \textbf{0.8220} & \textbf{0.6805} & \textbf{0.5445} & \textbf{0.4350} & \textbf{0.7768} & \textbf{0.5279} \\
		& NextMem-Sparse & \underline{0.8140}  & \underline{0.6740}  & \underline{0.5428}  & \underline{0.4326}  & \underline{0.7695}  & \underline{0.5244}  \\
		& \textcolor[rgb]{ .502,  .502,  .502}{*BGE} & \textcolor[rgb]{ .502,  .502,  .502}{0.8960} & \textcolor[rgb]{ .502,  .502,  .502}{0.7958} & \textcolor[rgb]{ .502,  .502,  .502}{0.6934} & \textcolor[rgb]{ .502,  .502,  .502}{0.6037} & \textcolor[rgb]{ .502,  .502,  .502}{0.9876} & \textcolor[rgb]{ .502,  .502,  .502}{0.6793} \bigstrut[b]\\
		\hline
		\hline
	\end{tabular}%
	\label{tab:result_task3}
	\vspace{-0.25cm}
\end{table*}

\textbf{Task 2: Contextual Generation (Memory Utilization)}\\
For latent memory, beyond preserving fine-grained details, it is crucial that the stored information can be utilized by LLMs for their inference.
Therefore, we evaluate our framework on memory utilization.
Specifically, for each query, we extract its references from datasets, and encode them into latent representations for LLMs to generate responses.
We design two settings: \textbf{(1) Compression (Comp.)}, where the model performs inference directly using latent representations, and \textbf{(2) DeCompression (DeComp.)}, where the inference is based on reconstructed information.
Besides the above baselines, we incorporate raw textual memory as an oracle comparison.
The results are presented in \textbf{Table~\ref{tab:result_task2}}.

According to the results, while ICAE shows an advantage in the Comp. settings, our models outperform all baselines in the DeComp. setting.
It indicates that while NextMem's latent space is less optimized for direct utilization, its superior reconstruction fidelity allows it to provide highly usable information once decompressed.
It also reveals a trade-off between reconstruction accuracy and instruction-following capability, which will be our next research topic in the future.
In contrast, DyPRAG and DeepSeek-OCR struggle to support effective generation in either setting.

\textbf{Task 3: Dense Passage Retrieval (Memory Retrieval)}\\
For most agent applications, maintaining long-term memory is essential to support inference at any future point, which makes the retrieval of query-relevant memory a key problem.
Since latent representation inherently possesses computational properties within the latent space, it is compatible as a retrieval index.
Therefore, we further evaluate their performances in retrieval.
Specifically, we generate latent representations for documents, pool them into 1D embeddings, and rank them based on matching scores calculated by cosine similarity with query embeddings.
We also incorporate BGE as a reference, despite its inability to reconstruct.
From the results in \textbf{Table~\ref{tab:result_task3}}, our methods demonstrate substantial improvements over other baselines.
In addition, NextMem also bridges the gap between latent memory and retrieval index.
These results indicate that NextMem can effectively unify memory storage and retrieval into a single latent representation, reducing the architectural complexity.

\subsection{Ablation Studies}
\label{sec:exp_ablation}
To evaluate the contribution of each component, we conduct an ablation study on RACE with the following settings: \textbf{(1) w/o ST}, removing \texttt{[SoD]} in inference. \textbf{(2) w/o PT}, removing the progressive latent substitution training. \textbf{(3) w/o PS}, excluding the progressive expansion in latent substitution training.
The results in \textbf{Table~\ref{tab:result_ablation}} shows the removal of any module leads to a performance degradation, with progressive latent substitution being the most critical.
In addition, \texttt{[SoD]} also plays a vital role, as its absence significantly drops the performance.
Moreover, the result shows that the progressive strategy can greatly improve the performance.
For the sparse model, the removal of scaling in quantization leads to a drastic decline.
These results indicate that each proposed component is important to latent memory storage.

\begin{table}[t]
	\centering
	\caption{Results of ablation study. The best results are highlighted in \textbf{bold}, and the second-best results are \underline{underlined}.}
	\vspace{-0.1cm}
	\resizebox{\linewidth}{!}
	{
	\begin{tabular}{>{\centering\arraybackslash}p{1.05cm}>{\centering\arraybackslash}p{1.05cm}>{\centering\arraybackslash}p{1.65cm}>{\centering\arraybackslash}p{1.35cm}>{\centering\arraybackslash}p{1.3cm}}
		\hline
		\hline
		\textbf{Methods} & \textbf{F1} & \textbf{ROUGE-L} & \textbf{METEOR} & \textbf{BertScore} \bigstrut\\
		\hline
		Dense & \textbf{0.8552} & \textbf{0.8580} & \textbf{0.8691} & \textbf{0.9735} \bigstrut[t]\\
		w/o ST & 0.3799  & 0.3804  & 0.4048  & 0.7307  \\
		w/o PT & 0.0159  & 0.0138  & 0.0169  & 0.7686  \\
		w/o PS & \underline{0.7389}  & \underline{0.7358}  & \underline{0.7353}  & \underline{0.9502}  \bigstrut[b]\\
		\hline
		Sparse & \textbf{0.8554} & \textbf{0.8583} & \textbf{0.8705} & \textbf{0.9731} \bigstrut[t]\\
		w/o SQ & 0.0309  & 0.0290  & 0.0442  & 0.7521  \bigstrut[b]\\
		\hline
		\hline
	\end{tabular}%
	}
	\label{tab:result_ablation}%
	\vspace{-0.6cm}
\end{table}%

\subsection{Influence of Compression Ratio Scaling}
\begin{figure}[t]
	\centering
	\begin{subfigure}[t]{\linewidth}
		\centering
		\includegraphics[width=\textwidth]{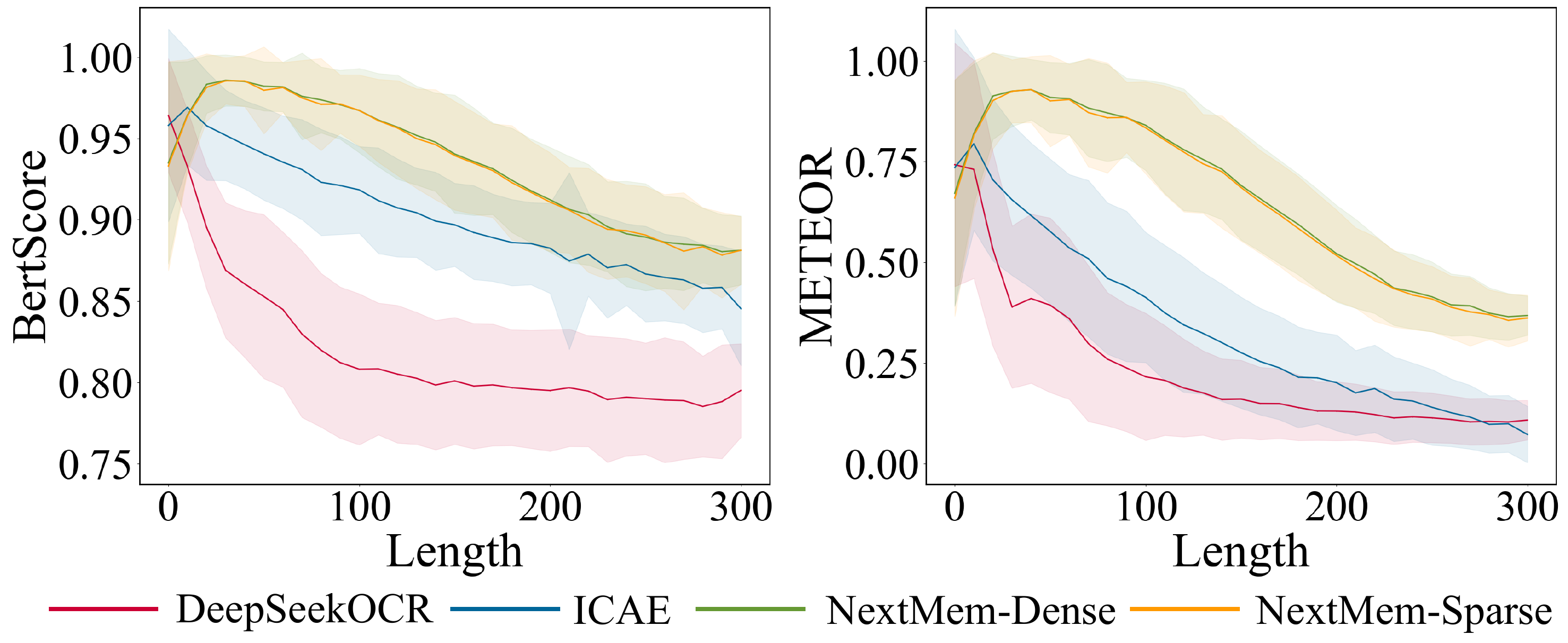}
	\end{subfigure}
	\vspace{-0.6cm}
	\caption{Results under varying compression ratios.}
	\vspace{-0.4cm}
	\label{fig:result_compression_ratio}
\end{figure}

Since the capacity of latent representations is intuitively limited, we further explore the model's scaling behavior in encoding and reconstruction across different text lengths.
The results are illustrated in \textbf{Figure~\ref{fig:result_compression_ratio}}.
We find that NextMem maintains higher performance as the input length increases comparing with other models.
While all models exhibit performance decay, NextMem shows a much slower and more graceful degradation.
In addition, we observe that NextMem has slight performance dips on shorter sequences, possibly due to hallucinations. 
Crucially, our model can maintain high semantic integrity beyond the training length (240 tokens), demonstrating robust extrapolation generalization to out-of-distribution sequence lengths.

\subsection{Robustness of Latent Memory}
\label{sec:exp_robustness}
\begin{figure}[t]
	\centering
	\begin{subfigure}[t]{1.0\linewidth}
		\centering
		\includegraphics[width=0.98\textwidth]{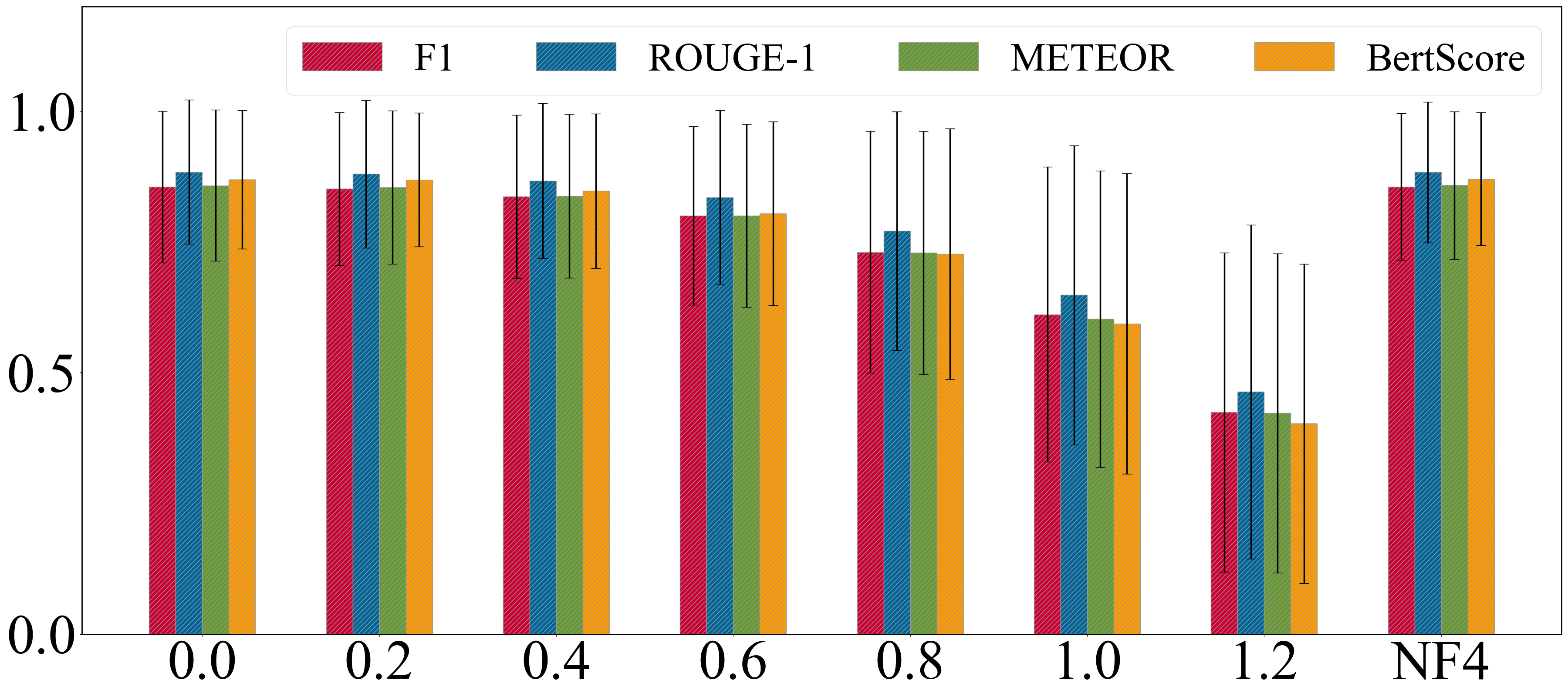}
	\end{subfigure}
	\vspace{-0.6cm}
	\caption{Robustness results of latent representations under varying levels of Gaussian noise ($\sigma$) and NF4 quantization.}
	\vspace{-0.6cm}
	\label{fig:result_robustness}
\end{figure}

Since latent memory may have precision loss or quantization to save storage, we further evaluate its robustness.
Specifically, we perturb the encoded latent memory by adding Gaussian noise $\epsilon \sim \mathcal{N}(0, \sigma^2)$.
From the results in \textbf{Figure~\ref{fig:result_robustness}}, we observe that our model maintains stable performance under moderate noise ($\sigma \le 0.8$), and still provides meaningful information under high perturbations.
Furthermore, the results show that NF4 quantization leads to negligible performance loss, which provides a foundation for NextMem-Sparse.
These findings validate that NextMem can effectively preserve information in noisy settings.

\subsection{Semantic Assignment Analysis}
\begin{figure}[t]
	\centering
	\begin{subfigure}[t]{\linewidth}
		\centering
		\includegraphics[width=\textwidth]{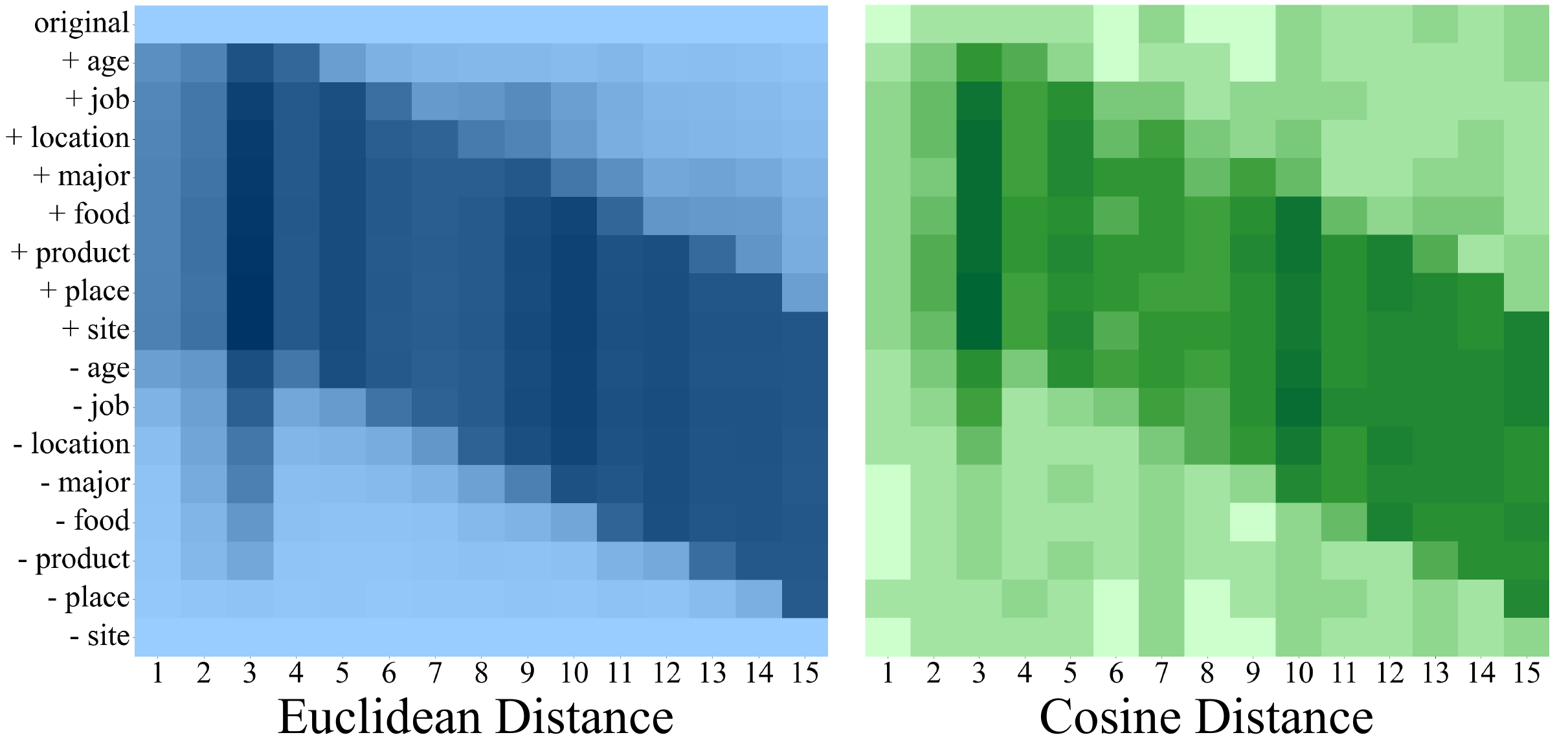}
	\end{subfigure}
	\vspace{-0.6cm}
	\caption{Semantic assignment analysis of latent memory.}
	\vspace{-0.4cm}
	\label{fig:result_causal}
\end{figure}

To explore the semantic assignment of latent memory, we perform an experiment using a paragraph with eight declarative sentences.
We iteratively substitute entities from the first sentence to the last, and then revert them.
For each step, we compute the distances between perturbed and original representations. 
Results in \textbf{Figure~\ref{fig:result_causal}} reveal a distinct diagonal pattern, which indicates a strong spatial-semantic mapping within latent memory.
It means specific memory positions are causally responsible for storing information from corresponding parts.
This relationship demonstrates that our framework successfully learns an ordered latent space, which is crucial for fine-grained memory editing.

\subsection{Training Procedure Analysis}
\begin{figure}[t]
	\centering
	\begin{subfigure}[b]{1.0\linewidth}
		\centering
		\includegraphics[width=\textwidth]{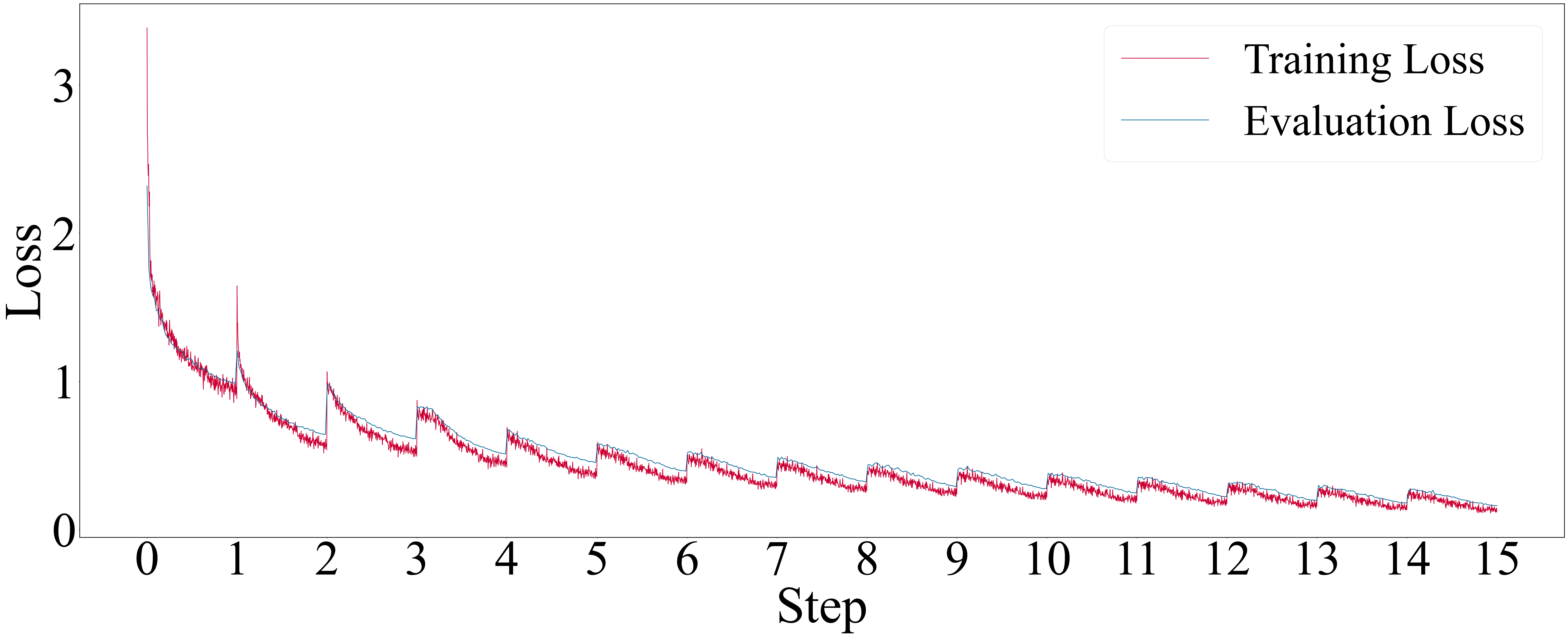}
	\end{subfigure}
	\vspace{-0.6cm}
	\caption{Curves of training and evaluation loss.}
	\vspace{-0.7cm}
	\label{fig:result_loss}
\end{figure}

\textbf{Figure~\ref{fig:result_loss}} presents the curves of training and evaluation losses, providing further insights into the model's optimization process.
Both losses exhibit downward trends throughout the training procedure, eventually converging to low values.
In addition, the loss curves display a distinct periodic sawtooth pattern that results from a progressive training strategy.
Temporary loss spikes occur at the boundaries where new substitution phases are introduced.
These spikes are immediately followed by rapid adaptation and continued loss reduction within each step.

\section{Related Works}
\label{sec:related_works}
Memory is important for LLM-based agents to store previous observations and for future inference, typically represented as textual and parametric forms~\cite{zhang2025survey}.
For textual memory, MemoryBank~\cite{zhong2024memorybank} stores conversations and summaries for the agent's reasoning.
MemGPT~\cite{packer2023memgpt} constructs an operating system to manage context memory.
For parametric memory, model editing methods can modify model parameters to inject knowledge~\cite{zhang2024comprehensive}.
\citet{zhang2025explicit} also explores their advantages and limitations across different tasks.
Some recent works try to employ latent tokens as experience for task learning and context management.
For example, Gist~\cite{mu2023learning} generates latent tokens from prefix task instructions for subsequent tasks.
TokMem~\cite{mu2023learning} constructs a trainable memory matrix to convert common instructions into embeddings.
MemGen~\cite{zhang2025memgen} integrates previous tokens to obtain implicit representations for current steps.


\section{Conclusion}
\label{sec:conclusion}
In this paper, we introduce NextMem, an autoregressive autoencoder framework for efficient latent factual memory.
By employing a two-stage training process and NF4 quantization, NextMem achieves high-fidelity reconstruction, robust retrieval, and significant storage reduction.
Experiments validate its effectiveness, providing a scalable and effective foundation for memory in LLM-based agents.

\clearpage

\section*{Impact Statements}
This paper presents work whose goal is to advance the field of machine learning. There are many potential societal consequences of our work, none of which we feel must be specifically highlighted here.


\bibliography{reference}
\bibliographystyle{icml2026}

\newpage
\appendix
\onecolumn

\section{Failure Cases}
\label{appendix:failure_case}

\subsection{Failure Cases on Reconstruction}
\textbf{Version 1 (Weighted Logits Combination):} Initially, rather than using hidden states, we derive the latent representation by weighting the input embeddings of the $\alpha$-quantile tokens based on their LM head logits at first.
This approach compresses memory while maintaining high sparsity, effectively lowering storage costs.
We use four special tokens, where \texttt{[SoD]/[EoD]} represent the start/end of textual documents, and \texttt{[SoM]/[EoM]} indicate the start/end of latent memory, respectively.
Our preliminary experiments involve direct training, without reconstruction alignment and progressive steps.
This method fails to yield the desired outcomes. While the loss drops from $2.18$ to $1.24$, the model merely learns to copy the reference text.
We suppose that it is hard to use existing sparse tokens to directly represent latent representations.

\textbf{Version 2 (Additional Latent Dictionary and Projection Head):}
To improve upon Version 1, we incorporate an extra latent representation dictionary, alongside an MLP projection head that maps hidden states to this dictionary.
These two parts are introduced as optimizable parameters.
In addition, we initialize the dictionary using PCA-reduced input embeddings followed by K-means clustering.
Therefore, the latent memory is combined with the tokens from the latent dictionary with projection logits.
Despite a loss reduction from $2.4$ to $1.7$, the model generates only meaningless text. We also find that the latent representations are very similar, indicating poor training of the latent space.
Moreover, we find the special tokens \texttt{[SoD]}, \texttt{[SoM]}, and \texttt{[EoD}] are not adequately learned. Given the requirement for training efficiency, we move away from this approach and pursue subsequent improvements.

\textbf{Version 3 (Two-Stage Training Strategy with Additional Latent Space):}
In this version, we introduce a two-stage training approach: autoregressive reconstruction alignment followed by latent space optimization.
The first stage focuses on explicit-to-explicit text replication, and the second stage establishes the process of encoding explicit text into latent representations and decoding them back to their original form.
We further streamline the model by discarding redundant special tokens from Version 2, using only \texttt{[SoD]} as a separator. While stage one performs successfully, which demonstrates reliable text replication, the loss of stage two stabilizes at around $2.8$, showing no meaningful convergence.
We suspect it may result from the fact that LM head maps high-dimensional hidden state onto discrete tokens, losing semantic information.
We also suspect that it is hard to train all latent tokens, so they should be trained in a progressive way.

\textbf{Version 4/Final (Progressive Latent Substitution Strategy):}
In this version, we implement two critical modifications. First, we introduce progressive latent substitution, which replaces the previous one-time optimization approach with a gradual substitution strategy.
Second, we transition from the weighted combination of logits to the direct generation of latent hidden states. These changes result in a stable and persistent decline in training loss.
Furthermore, based on experiments, we incorporate several key optimizations:
(1) We append an \texttt{<end\_of\_text>} token to explicitly denote the completion of the decoding process.
(2) We increase the training to three epochs, ensuring the loss reaches a sufficiently low and appropriate level following the substitution-based augmentation.
After performing hyperparameter tuning, such as learning rate, our final models are successfully produced by this approach, with stable and reliable encoding and decoding capability.

\subsection{Failure Cases on Sparsity}
\textbf{Version 5 (Mixture-of-Experts Strategy):}
To further minimize the storage overhead of latent memory, we explore sparsification techniques inspired by the Mixture-of-Experts (MoE) framework. We implement a gating function to manage individual experts, which we construct as a mapping codebook initialized via Singular Value Decomposition (SVD) of the input embeddings.
In this setup, we treat the hidden states as queries to perform sparsification and approximate reconstruction, optimizing the entire pipeline during the progressive latent substitution phase.
However, experimental results show that the training loss remains stagnant at approximately $3.3$, failing to converge further.

\textbf{Version 6 (RQ-VAE Strategy):}
We further try RQ-VAE~\cite{rajput2023recommender} framework to achieve latent representation sparsification.
It discretizes continuous inputs into a hierarchical sequence of codebook indices through a recursive quantization process. Initially, the input is mapped into a latent embedding space.
This embedding is then iteratively decomposed across multiple levels.
At each stage, the model identifies the nearest codeword from a level-specific codebook based on Euclidean distance and calculates a residual to be passed to the subsequent level.
The final latent representation is reconstructed by aggregating the selected codewords from all levels through element-wise summation.
However, experimental results show that while the training loss decreases to approximately $1.0$, the model encounters a representation collapse, where latent vectors across all positions become nearly identical.
Additionally, the decoded outputs exhibit significant overfitting, primarily reflecting memorized training data rather than generalized features.

\textbf{Version 7 (OMP Strategy):}
We further employ the Orthogonal Matching Pursuit (OMP) strategy to sparsify the generated latent representations.
For each latent vector, OMP iteratively selects the atoms from a pre-defined dictionary that exhibit the highest correlation with the current residual, followed by an orthogonal projection to optimize the corresponding coefficients.
The signal is then reconstructed as a sparse linear combination of these selected atoms, with the fidelity of the approximation evaluated by the L2 norm of the residual between the original and reconstructed vectors.
However, experimental results reveal significant residuals, indicating that the OMP process fails to achieve an accurate approximation of the original signals. This high reconstruction error suggests that the latent space may not sufficiently match the original representations.

\textbf{Version 8 (Explicit Projection):}
In this version, we explore explicit projection by tasking the trained decoder with translating latent space vectors into their respective logits and explicit tokens. We expect that the model can effectively summarize the original explicit text into a single token through this process. However, the model merely selects isolated tokens from the original sequence in a disjointed pattern. It fails to capture the cohesive, compressed semantic information of the entire text segment, instead focusing on individual token-level mappings.

\textbf{Version 9 (Reparameterization Strategy):}
Furthermore, we employ Gumbel-Softmax and Gaussian Softmax reparameterization tricks to bridge the gap between discrete token generation and subsequent computations.
This approach aims to address the non-differentiability of discrete variables, allowing gradients to flow through the sampling process. However, we find that this strategy leads to significant numerical instability. The training loss fluctuates between $1.0$ and $2.0$, which leads to poor performance and increases decoding hallucinations in the generated output.

\section{Details of Reproduction}
\label{appendix:reproduction}
In this section, we describe the details of the model and experiment reproduction.

\subsection{Dataset Preparation}

\textbf{Overview.} For each dataset, we standardize the data into a unified format consisting of three primary fields, including the question, answer, and reference.
The processing pipeline for each dataset is outlined as follows: 

\textbf{$\bullet$ HotpotQA\footnote{\url{http://curtis.ml.cmu.edu/datasets/hotpot/hotpot_train_v1.1.json}}:}
We extract \texttt{easy} level samples to focus on direct evidence retrieval. For each instance, we reconstruct the reference list by extracting only the specific sentences identified as \texttt{supporting facts} from the original context, rather than using the entire document. The final dataset is divided into a 90\% for training and a 10\% for testing.\\
\textbf{$\bullet$ RACE\footnote{\url{http://www.cs.cmu.edu/~glai1/data/race/RACE.tar.gz}}:}
We aggregate the \texttt{High} level data across the original datasets.
We consider the context of \texttt{article} as the reference. 
We then re-partition the combined pool into 90\% for training and 10\% for testing.\\
\textbf{$\bullet$ SQuAD\footnote{\url{https://rajpurkar.github.io/SQuAD-explorer/dataset/train-v2.0.json}}:} To adapt it for the short-term memory task, we filter out all unanswerable questions, retaining only those where the answer is explicitly present in the context. The original context is utilized as the reference, and the various ground-truth answers are compiled into a list. The processed data is partitioned into a training set and a test set with 9:1.\\
\textbf{$\bullet$ LoCoMo\footnote{\url{https://github.com/snap-research/LoCoMo}}:}
From the LoCoMo dataset, we select samples belonging to category 1 (\textit{i.e.,} \texttt{multi-hop}) and category 5 (\textit{i.e.,} \texttt{single-hop}), which require specific evidence-based reasoning. We map the \texttt{evidence} IDs provided in the original metadata to the corresponding speaker-text turns in the movie dialogues to form the reference list. This dataset serves as a specialized test suite for long-context conversation understanding.\\
\textbf{$\bullet$ LongMemEval\footnote{\url{https://huggingface.co/datasets/xiaowu0162/longmemeval-cleaned}}:}
We utilize a cleaned version of the LongMemEval-S dataset for evaluation. The references for each question are constructed by aggregating relevant dialogue turns from multiple \texttt{haystack} sessions. Unlike other datasets, this set is primarily used for testing and evaluation of memory retrieval capabilities across long-context dialogues.

In addition, the procedure for processing the evaluation data for each task is as follows:

\textbf{Task 1: Factual Reconstruction.}
We further construct a reconstructed reference pool derived from the datasets.
To manage the data scale and simulate varying context lengths, the script implements a two-stage refinement strategy.
First, it performs strided sampling based on a dataset-specific sampling gap to select a representative subset of the data.
Second, for every reference associated with the sampled entries, a stochastic truncation is applied: the text is split into sentences, and only the first $n$ sentences (where $n$ is randomly sampled from $[1, 8]$) are retained.
The resulting fragments are normalized to ensure proper punctuation and are finally saved as a flattened list of strings. This procedure effectively creates a diversified lengths of reference snippets for evaluating memory reconstruction capabilities.\\
\textbf{Task 2: Contextual Generation.}
We utilize the processed testing data for the evaluation of contextual generation.

\textbf{Task 3: Dense Passage Retrieval.}
Compared with the initial processing, we further construct the datasets for evaluating memory retrieval evaluation by introducing a \texttt{hit\_list}, which records the precise indices of the ground-truth references within a larger context pool.
While the first process primarily paired questions with their necessary evidence, this version expands the references field to include non-essential or session-wide context and explicitly tracks which specific entries are required to answer the question.
It allows a more rigorous evaluation of a model's ability to accurately retrieve.

\subsection{Model and Training Configuration}
For the autoregressive autoencoder, we employ Qwen3-8B\footnote{\url{https://huggingface.co/Qwen/Qwen3-8B}} as the backbone model.
The maximum lengths for both encoding and output are set to 1024 tokens.
During the training phase, the progressive training scheme consists of 15 steps with a block size of 16.
Parameter-efficient fine-tuning is performed via LoRA, where we set the rank $r=16$, $\alpha=32$, and a dropout rate of $0.1$.
We apply LoRA adapters to the \texttt{q\_proj}, \texttt{k\_proj}, \texttt{v\_proj}, and \texttt{o\_proj} modules.
The model is trained on 4 NVIDIA A100 GPUs for 3 epochs per step, with an effective batch size of $32$ and a learning rate of $5 \times 10^{-4}$.

\subsection{Baseline Configuration}
\textbf{DeepSeek-OCR.}
For DeepSeek-OCR, we employ the PIL\footnote{\url{https://pypi.org/project/pillow/}} package to render raw text into images, enabling the model to perform decoding.
To align with the model’s latent token capacity, the images are generated at a resolution of $240 \times 240$ pixels, which corresponds to 16 latent tokens.
Specifically, the text is rendered in black on a white background using the Times New Roman font (size 12), with a 5-pixel padding and an automatic line-wrap mechanism.
Following the original model configuration~\footnote{\url{https://github.com/deepseek-ai/DeepSeek-OCR/}}, we use the decoding prompt: \texttt{<image>\textbackslash nOCR this image}.

\textbf{ICAE.}
For ICAE, we utilize the official checkpoint \texttt{mistral\_7b\_ft\_icae}\footnote{\url{https://huggingface.co/sggetao/icae}} based on the Mistral-7B-Instruct-v0.2\footnote{\url{https://huggingface.co/mistralai/Mistral-7B-Instruct-v0.2}} to generates 128 latent tokens. Its LoRA fine-tuning configuration consists of a rank $r=512$ and a dropout rate of 0.05. 

\textbf{DyPRAG.}
For DyPRAG, we employ checkpoint \texttt{llama3-8b-p32-1ep-main-2400sample}, built upon Llama-3-8B\footnote{\url{https://huggingface.co/meta-llama/Meta-Llama-3-8B-Instruct}} to generate LoRA adapters. The LoRA parameters are set to a rank $r=2$ and $\alpha=32$, with the projector $p=32$.

\textbf{Textual Memory.}
This model is exclusively employed in Task 2 for a comparison of contextual utilization, using prompt: \\
\texttt{
Please answer the following question based on the reference.\\
Reference:\\
\{reference\}\\ \\
Question:\\
\{question\}\\ \\
Answer:}

\textbf{BGE.}
This model is exclusively utilized in Task 3 for a comparison of dense passage retrieval performance. Specifically, we employ BGE-M3~\footnote{\url{https://huggingface.co/BAAI/bge-m3}}, loaded via SentenceTransformers\footnote{\url{https://pypi.org/project/sentence-transformers}}.

\section{Influence of Block Size}
\label{appendix:exp_block_mask_size}

We further explore how the substitution block size and the final length affect the performance, which are both significant hyper-parameters in our framework. The results are presented in \textbf{Figure~\ref{fig:result_extensive}(a)}.
We observe a clear positive correlation between performance and latent length $L$, as increasing the memory capacity allows for more nuanced information storage. Furthermore, the block size $B$ balances efficiency and accuracy.
When $B$ is small (e.g., $B=8$), the model struggles to maintain reconstruction for long paragraphs given a certain latent length.
We also find that a large $B$ can also be suboptimal for optimization, with performance peaking at $B=16$.
This is consistent with our findings in \textbf{Section~\ref{sec:exp_ablation}} (w/o PS), suggesting that optimizing long tokens in a single stage is inherently difficult.

\section{Forgetting Effect of Latent Memory}
\label{appendix:forgetting_effect}
\begin{figure}[t]
	\centering
	
	\begin{subfigure}[b]{0.53\linewidth}
		\centering
		\includegraphics[width=\textwidth]{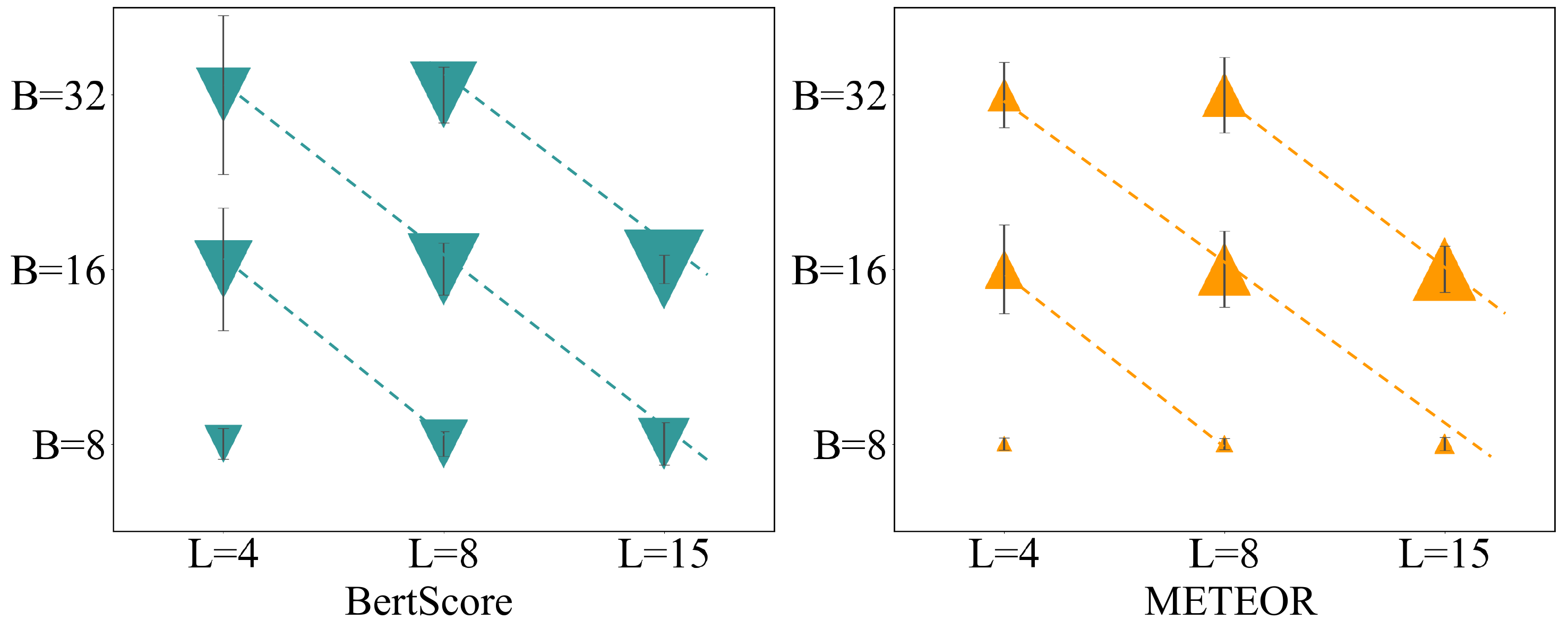}
		\subcaption{Results of different block sizes and latent lengths.}
		\vspace{-0.1cm}
	\end{subfigure}
    \begin{subfigure}[b]{0.385\linewidth}
		\centering
		\includegraphics[width=\textwidth]{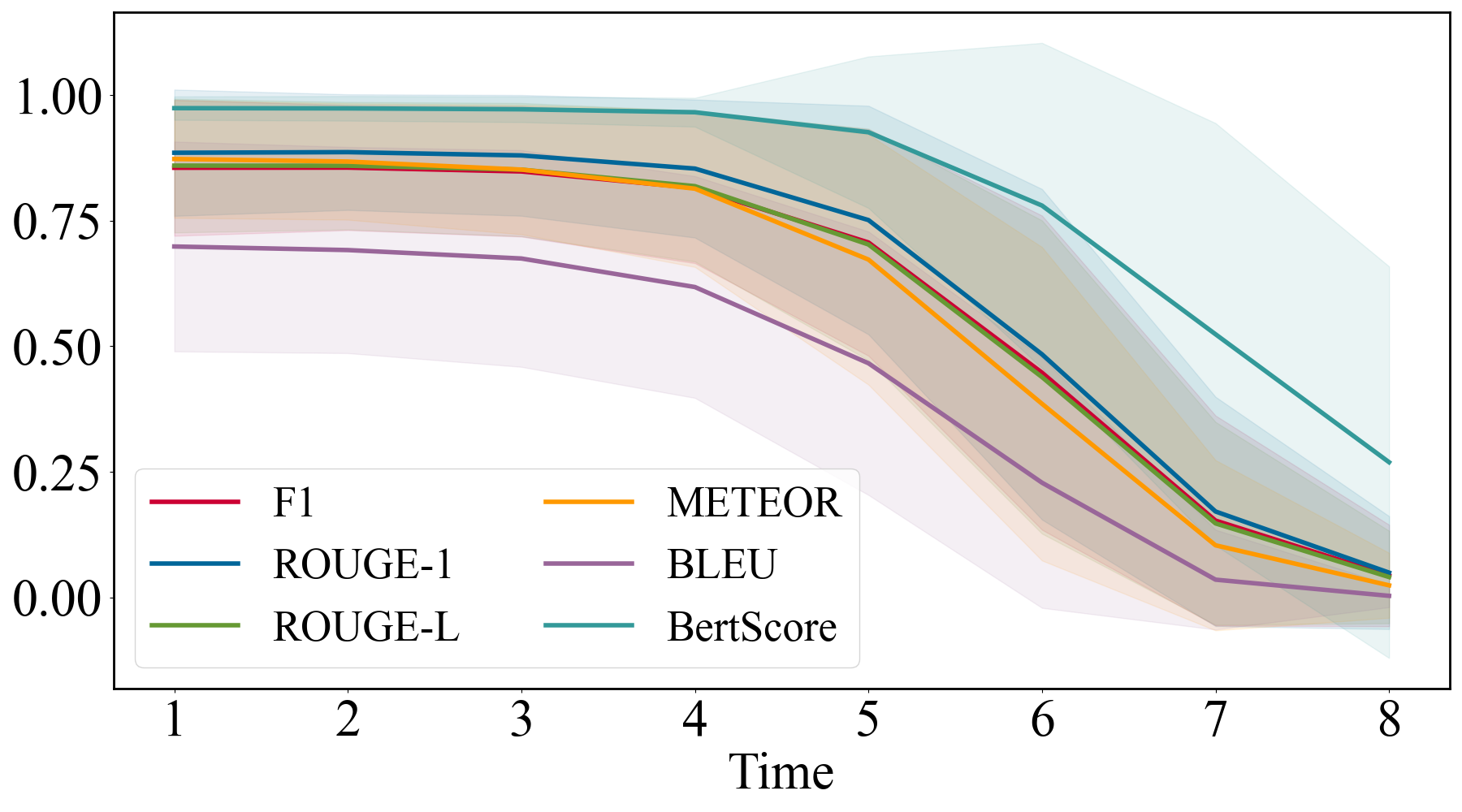}
		\subcaption{Results of forgetting effects.}
		\vspace{-0.1cm}
	\end{subfigure}
	\caption{Results of extensive experiments.}
	\vspace{-0.5cm}
	\label{fig:result_extensive}
\end{figure}

Furthermore, we simulate the memory decay process by introducing a pooling coefficient $\alpha(t) = a^t$, where $a$ controls the rate of forgetting. 
At each time step $t$, the pooled representation is $\mathbf{H}^{(L)} = \mathbf{H}^{(L)} \cdot \alpha(t) + \bar{\mathbf{H}}^{(L)} \cdot (1-\alpha(t))$, where $\bar{\mathbf{H}}^{(L)}$ is the average embedding across various lengths.
The results are presented in \textbf{Figure~\ref{fig:result_extensive}(b)}.
We find the performance across all evaluation metrics remains relatively stable during the initial time steps ($t \leq 4$).
However, as $t$ increases further, we observe a sharp and consistent decline in all scores.
This downward trend demonstrates that as the coefficient $\alpha(t)$ decreases, the latent memory effectively forgets specific factual details by gradually converging toward the average representation $\bar{\mathbf{H}}^{(L)}$.

\clearpage

\section{Case Studies}
\label{appendix:case_study}

\subsection{Case Study on Paragraphs with Various Lengths}
\textbf{Case Study: Below Length Limits.}\\
At this length, the model demonstrates near-perfect reconstruction fidelity. The semantic content, syntactic structure, and specific entities are almost identical to the original text. The reconstruction preserves every detail, from Alice's specific job title in Silicon Valley to her dinner preferences. Minor variations are negligible (\textit{e.g.,} capitalizing Master's), illustrating that within this limit, the model can effectively store and retrieve the full information density of the paragraph without loss.

\begin{figure*}[h]
	\centering
	\vspace{-0.1cm}
	\begin{tcolorbox}[
		colback=SoftBlue!50, 
		colframe=SteelBlue, 
		coltitle=SteelBlue,
		arc=0.05cm,
		boxrule=0.8pt,
		]
		\textbf{Original Paragraphs (91 words):}
		Alice recently celebrated her 21st birthday with a memorable rooftop party surrounded by her closest friends. She works as a specialized software engineer for a fast-paced technology startup in Silicon Valley. She currently resides in a cozy, sun-drenched apartment in the heart of downtown Los Angeles. She holds a master's degree in Computer Science, having graduated with honors from a prestigious university. She prefers authentic Italian pasta with fresh basil and creamy tomato sauce for dinner. She bought a new high-performance laptop yesterday afternoon to help with her freelance coding projects.\\
		\textbf{Reconstruction:}
		Alice celebrated her 21st birthday with a memorable rooftop party surrounded by her closest friends. She works as a specialized software engineer for a fast-paced technology startup in Silicon Valley. She currently resides in a cozy, sun-drenched apartment in the heart of downtown Los Angeles. She holds a Master's degree in Computer Science, having graduated with honors from a prestigious university. She prefers authentic Italian pasta with fresh basil and creamy tomato sauce for dinner. She bought a new high-performance laptop yesterday afternoon to help with her freelance coding projects.
	\end{tcolorbox}
	\vspace{-0.1cm}
\end{figure*}

\textbf{Case Study: At Length Limits.}\\
As the text approaches the length limit, the model demonstrates impressive semantic robustness. It successfully preserves the core narrative structure and the vast majority of the character's profile attributes.
While minor associative substitutions occur, the model maintains the overall logical coherence and thematic consistency of the original text. For example, it replaces specific models (\textit{i.e.,} Tesla Model 3 to Model S) or instruments (\textit{i.e.,} violin to bass). This indicates that even at full capacity, the model effectively prioritizes the retention of high-level biographical information and functional context.

\begin{figure*}[h]
	\centering
	\vspace{-0.1cm}
	\begin{tcolorbox}[
		colback=SoftBlue!50, 
		colframe=SteelBlue, 
		coltitle=SteelBlue, 
		arc=0.05cm,
		boxrule=0.8pt,
		]
		\textbf{Original Paragraphs (190 words):}
		Alice recently celebrated her 21st birthday with a memorable rooftop party surrounded by her closest friends. She works as a specialized software engineer for a fast-paced technology startup in Silicon Valley. She currently resides in a cozy, sun-drenched apartment in the heart of downtown Los Angeles. She holds a master's degree in Computer Science, having graduated with honors from a prestigious university. She prefers authentic Italian pasta with fresh basil and creamy tomato sauce for dinner. She bought a new high-performance laptop yesterday afternoon to help with her freelance coding projects. She visited the interactive science museum last weekend to explore the new exhibition on space travel. Alice will travel to Spain next week to spend ten days exploring the historic streets of Barcelona and Madrid. She plays the electric violin in a local jazz band that performs at community events every month. She drives a sleek, cherry-red Tesla Model 3 that she carefully maintains in pristine condition. She speaks fluent Japanese and English, a skill she developed while growing up in a bilingual household. She adopted a fluffy white cat with bright blue eyes from the animal shelter today.
		\\
		\textbf{Reconstruction:}
		Alice celebrated her 21st birthday with a memorable rooftop party surrounded by her closest friends. She worked as a specialized software engineer for a tech startup in Silicon Valley. She currently resides in a cozy, sun-drenched downtown apartment in San Francisco. She holds a Master's degree in Computer Science, with honors, from a prestigious university. She prefers authentic Italian pasta with fresh creamy tomato sauce and basil for dinner. She bought a new high-performance laptop yesterday afternoon to help with her freelance coding assignments. She visited the international space exploration lab last Saturday to explore the new interactive display. She will spend the next two weeks in Spain traveling to explore the historic streets of Madrid and Barcelona. She plays the electric bass in a local music club that performs popular songs in Spanish and English. She drives a sleek, cherry-red Tesla Model S that she maintains in pristine condition. She cherishes the skill she learned in Spanish and English, a language she developed while growing up in a bilingual household. She adopted a fluffy white bunny from the animal shelter with bright blue eyes today.	
	\end{tcolorbox}
	\vspace{-0.1cm}
\end{figure*}

\clearpage

\textbf{Case Study: Over Length Limits.}\\
Even when significantly exceeding the designated length limits, the model exhibits a remarkable ability to reconstruct the overarching primary life events. Key elements such as the subject's professional background, academic honors, and upcoming travel plans remain intact. Although the extreme information density leads to some blending of secondary details, the model still manages to capture the essential persona of the original text. This resilience suggests that the model’s memory mechanism can maintain a coherent thematic framework even when pushed well beyond its operational window.

\begin{figure*}[h]
	\centering
	\begin{tcolorbox}[
		colback=SoftBlue!50, 
		colframe=SteelBlue, 
		coltitle=SteelBlue, 
		arc=0.05cm,
		boxrule=0.8pt,
		]
		\textbf{Original Paragraphs (263 words):}
		Alice recently celebrated her 21st birthday with a memorable rooftop party surrounded by her closest friends. She works as a specialized software engineer for a fast-paced technology startup in Silicon Valley. She currently resides in a cozy, sun-drenched apartment in the heart of downtown Los Angeles. She holds a master's degree in Computer Science, having graduated with honors from a prestigious university. She prefers authentic Italian pasta with fresh basil and creamy tomato sauce for dinner. She bought a new high-performance laptop yesterday afternoon to help with her freelance coding projects. She visited the interactive science museum last weekend to explore the new exhibition on space travel. Alice will travel to Spain next week to spend ten days exploring the historic streets of Barcelona and Madrid. She plays the electric violin in a local jazz band that performs at community events every month. She drives a sleek, cherry-red Tesla Model 3 that she carefully maintains in pristine condition. She speaks fluent Japanese and English, a skill she developed while growing up in a bilingual household. She adopted a fluffy white cat with bright blue eyes from the animal shelter today. She attended an energetic outdoor music concert last Friday night despite the rainy weather. She drinks a large iced coffee without sugar every morning to boost her energy for the day. She reads gripping mystery novels by Agatha Christie for at least an hour before sleeping. Alice will move to a more spacious apartment in a quieter neighborhood soon. She is severely allergic to roasted peanuts and always checks ingredients carefully before eating.\\
		\textbf{Reconstruction:}
		Alice celebrated her 21st birthday with a memorable rooftop party surrounded by her closest friends. She worked as a specialized software engineer at a high-tech startup in Silicon Valley. She resides in a sun-drenched, cozy downtown apartment in the heart of San Francisco. She holds a Master's degree in Computer Science, with honors. She prefers authentic Italian pasta with fresh basil and creamy tomato sauce. She bought a new high-performance laptop yesterday afternoon to help with her freelance drawing work. She visited the interactive science center last week to explore the new space exploration module. Alice is going to Spain next week to spend ten days exploring the historic Spanish cities. She plays the electric violin and explores the local jazz clubs in Madrid. She enjoys a monthly party at a cherry-bright electric car that she drives in a sleek, pristine condition. She maintains a clean, sharp Mercedes E-Class she drives in a professional manner. She learned Spanish and fluent Japanese while growing up, a skill she developed in a bilingual environment. She adopted a fluffy white bunny from the animal shelter today with bright blue eyes. She enjoys an energetic outdoor workout despite the rainy evening. She drank a large iced coffee without sugar last night. She has no dietary restrictions.
	\end{tcolorbox}
\end{figure*}

\clearpage

\subsection{Case Study on Gaussian Noise}
The case study demonstrates a clear inverse relationship between the noise level ($\sigma$) and the semantic fidelity of the reconstructed text. The settings are aligned with \textbf{Section~\ref{sec:exp_robustness}}. The degradation process can be categorized into four levels:\\
\textbf{$\bullet$ High Fidelity ($\sigma = 0.4$):} At this level, the reconstruction is nearly identical to the original text. The model preserves all factual details (\textit{e.g.,} age, occupation, location, and degree) with only negligible omissions (\textit{e.g.,} the word recently), maintaining near-perfect consistency.\\
\textbf{$\bullet$ Minor Semantic Drift ($\sigma = 0.8$):} The text remains grammatically fluent, but factual hallucinations begin to surface. While the core narrative is intact, specific entities are substituted with contextually plausible but incorrect ones. For instance, changing the residence from Los Angeles to San Francisco. This indicates that moderate noise begins to interfere with the model's ability to retrieve precise tokens.\\
\textbf{$\bullet$ Structural Breakdown ($\sigma = 1.2$):} Consistency degrades significantly as the noise disrupts the logic of the sentence. The model starts to repeat irrelevant descriptors (\textit{e.g.,} the repetitive use of 10th-floor) and fails to complete the paragraph. The original meaning is partially lost, replaced by fragmented imagery.\\
\textbf{$\bullet$ Semantic Collapse ($\sigma \ge 1.6$):} At extreme noise levels, the model suffers from total divergence or output failure. The reconstruction either hallucinates entirely unrelated content (\textit{e.g.,} introducing a random date 2015-08-23 and changing the subject's gender to he) or fails to generate any text at all.

\begin{figure*}[h]
	\centering
	\begin{tcolorbox}[
		colback=SoftBlue!50, 
		colframe=SteelBlue, 
		coltitle=SteelBlue, 
		detach title,
		before upper={\textcolor{SteelBlue}{\textbf{Case Study on Gaussian Noise}}\medskip\hrule\medskip},
		arc=0.05cm,
		boxrule=0.8pt,
		]
		\textbf{Original Paragraphs (62 words):}
		Alice recently celebrated her 21st birthday with a memorable rooftop party surrounded by her closest friends. She works as a specialized software engineer for a fast-paced technology startup in Silicon Valley. She currently resides in a cozy, sun-drenched apartment in the heart of downtown Los Angeles. She holds a master's degree in Computer Science, having graduated with honors from a prestigious university. \\
		\textbf{Noisy Reconstruction ($\sigma=0.4$):}
		Alice celebrated her 21st birthday with a memorable rooftop party surrounded by her closest friends. She works as a specialized software engineer for a fast-paced technology startup in Silicon Valley. She currently resides in a cozy, sun-drenched apartment in the heart of downtown Los Angeles. She holds a Master's degree in Computer Science, having graduated with honors from a prestigious university. \\
		\textbf{Noisy Reconstruction ($\sigma=0.8$):}
		Alice celebrated her 21st birthday with a memorable rooftop party surrounded by her closest friends. She works as a specialized software engineer for a fast-paced technology startup in Silicon Valley. She currently resides in a cozy, sun-drenched apartment in the heart of downtown San Francisco. She holds a Master's degree in Computer Science, having graduated from a prestigious university. \\
		\textbf{Noisy Reconstruction ($\sigma=1.2$):}
		Alice celebrated her 21st birthday with a memorable weekend trip to the 10th-floor rooftop with her closest friends. She works as a specialized software engineer in a fast-paced 10th-floor startup. She currently resides in a cozy, sun-drenched, and dry \\
		\textbf{Noisy Reconstruction ($\sigma=1.6$):}
		2015-08-23: The world's most popular celebration of a warm summer night includes a group of close friends. He works as a specialized software engineer for a fast-paced technology company. He resides in a sun-drenched, cozy house in the center of the city.\\
		\textbf{Noisy Reconstruction ($\sigma=1.6$):} (just a single space)
	\end{tcolorbox}
\end{figure*}

\end{document}